%% file: dalton-arxiv.tex
\documentclass[twoside]{article}

\usepackage[preprint]{aistats2024_arxiv}
\usepackage[round]{natbib}

\usepackage{dalton}
\usepackage{xr}

\begin{document}
    
    % If your paper is accepted and the title of your paper is very long,
    % the style will print as headings an error message. Use the following
    % command to supply a shorter title of your paper so that it can be
    % used as headings.
    %
    %\runningtitle{I use this title instead because the last one was very long}
    
    % If your paper is accepted and the number of authors is large, the
    % style will print as headings an error message. Use the following
    % command to supply a shorter version of the authors names so that
    % they can be used as headings (for example, use only the surnames)
    %
    %\runningauthor{Surname 1, Surname 2, Surname 3, ...., Surname n}
    
    \twocolumn[
    
    \aistatstitle{Data-Adaptive Probabilistic Likelihood Approximation for Ordinary Differential Equations}
    
    \aistatsauthor{
        Mohan Wu 
        % examples of more authors
        \And
        Martin Lysy
    }
    
    \aistatsaddress{ 
        University of Waterloo
        \And
        University of Waterloo
    }]
    
    \input{main}
    \bibliography{rodeo-ref} 
    \appendix
    \onecolumn
    \input{supp}

\end{document}

%% file: main.tex
\begin{abstract}
  Estimating the parameters of ordinary differential equations (ODEs) is of fundamental importance in many scientific applications. 
  % Parameter inference for ordinary differential equations (ODEs) is of fundamental importance in many scientific applications.
  While ODEs are typically approximated with deterministic algorithms, new research on probabilistic solvers indicates that they produce more reliable parameter estimates by better accounting for numerical errors.  However, many ODE systems are highly sensitive to their parameter values.  This produces deep local maxima in the likelihood function -- a problem which existing probabilistic solvers have yet to resolve.
  Here we present a novel probabilistic ODE likelihood approximation, \dalton, which can dramatically reduce parameter sensitivity
  % sensitivity to model parameters
  by learning from noisy ODE measurements in a data-adaptive manner.
  % Here we present a novel probabilistic approximation to the ODE likelihood function which can dramatically reduce sensitivity to model parameters.  This is achieved by learning from noisy ODE measurements in a data-adaptive manner.
  Our approximation scales linearly in both ODE variables and time discretization points, and is applicable to ODEs with both partially-unobserved components and non-Gaussian measurement models.  Several examples demonstrate that \dalton produces more accurate parameter estimates via  numerical optimization than existing probabilistic ODE solvers, and even in some cases than the exact ODE likelihood itself. 
\end{abstract}

\section{Introduction}

Parameter estimation for ordinary differential equations (ODEs) is an important statistical and machine learning problem in the natural sciences and engineering.  Typically this is accomplished via parameter searching algorithms which repeatedly solve the ODE at successive evaluations of the likelihood function.  However, many ODE systems are hypersensitive to their input parameters, resulting in sharp local maxima in the likelihood function from which parameter search algorithms may fail to escape~\citep{kohberger78,cao11,dass17,rackauckas18,weber18}.

Since most ODEs do not have-closed form solutions, they must be approximated by numerical methods.  
% However, many ODE systems are hypersensitive to their input parameters, 
% leading numerical solvers to converge to the local minima~\citep{kohberger78,cao11,dass17,rackauckas18,weber18}.
While traditionally this has been done with deterministic algorithms~\citep[e.g.,][]{butcher08,griffiths.higham10,atkinson.etal09}, a growing body of work in \emph{probabilistic numerics}~\citep{diaconis88,skilling92,hennig.etal15} indicates that probabilistic ODE solvers, which directly account for uncertainty in the numerical approximation, provide more reliable parameter estimates in ODE learning problems~\citep{chkrebtii.etal16,conrad.etal17}.
% Traditionally this has been done with deterministic algorithms~\citep[e.g.,][]{butcher08,griffiths.higham10,atkinson.etal09}. However, a growing body of work in \emph{probabilistic numerics}~\citep{diaconis88,skilling92,hennig.etal15} indicates that probabilistic ODE solvers, which directly account for uncertainty in the numerical approximation, provide more reliable parameter estimates in ODE learning problems~\citep{chkrebtii.etal16,conrad.etal17}.
In particular, probabilistic solvers have the ability to condition on the observed data to guide the ODE solution, which can decrease sensitivity to model parameters~\citep{chkrebtii.etal16,yang.etal21,schmidt21,tronarp22}.  However, due to the increased complexity relative to deterministic solvers, the potential for probabilistic ODE solvers to reduce parameter
% of probabilistic solvers relative to their deterministic counterparts, their potential to reduce parameter
hypersensitivity in a computationally efficient manner has yet to be fully realized.

\paragraph{Contributions} Here we present a novel
% \underline{D}ata-\underline{A}daptive probabi\underline{L}is\underline{T}ic ODE likelih\underline{O}od approximatio\underline{N}
\underline{d}ata-\underline{a}daptive probabilistic ODE \underline{l}ikelihood approxima\underline{t}i\underline{on} (\dalton) which attempts to bridge this gap.  Several examples will serve to illustrate the key features of \dalton, which are:
%At the heart of our approach is a Bayesian filtering paradigm~\cite{tronarp.etal18} 
% making heavy use of the Kalman filtering and smoothing recursions~\citep{tronarp.etal18},
%particularly noted for its high accuracy and linear complexity in both time discretization points and the number of system variables~\citep[e.g.,][]{kramer21,bosch21,bosch22,tronarp22}. We show how to approximately condition on the observed data in the forward pass of this Bayesian filtering model, allowing for data which is both partially unobserved and subject to arbitrary non-Gaussian measurement errors. 
% For the purpose of ODE parameter learning from noisy observations, we present an approximation of the likelihood function via probabilistic numerics which:
\begin{itemize}
%\item \emph{Convergence}: Provides empirical evidence of convergence to the exact ODE likelihood as the discretization interval decreases.
\item \emph{Scalability:} Building upon a Bayesian filtering paradigm for probabilistic ODE solvers~\citep{tronarp.etal18} and associated approximation methods~\citep[e.g.,][]{tronarp.etal18, schober.etal19, kramer21}, the \dalton algorithm scales linearly with both the number of time discretization points and the number of ODE variables.
\item \emph{Flexibility:} \dalton can be applied to arbitrary order ODEs with both partially-observed components and non-Gaussian measurement models.  The usefulness of the latter is demonstrated on an application to infectious disease modeling (example~\ref{sec:seirah}).
\item \emph{Robustness:} \dalton is able to overcome deep local modes in the ODE likelihood function that existing probabilistic and deterministic solvers cannot.  This is demonstrated via insensitivity to initial values of a standard mode-finding algorithm for Bayesian estimation of the parameters of both oscillatory and chaotic ODE systems (examples~\ref{sec:osc} and~\ref{sec:lorenz}), for which \dalton is shown to be more reliable than using the true ODE likelihood itself.  
% \item \emph{Non-Gaussian Errors}: Generalizes to non-Gaussian errors, the usefulness of which is demonstrated on an application from infectious disease modeling.
% \item \emph{Partially Unobserved Data}: Applies to data with partially unobserved components.
\end{itemize}
% Several examples illustrate that \dalton is more accurate than leading probabilistic ODE parameter learning methods, and in the extreme case of chaotic ODE systems, is more reliable even than using the true ODE likelihood itself. 

\section{Background}\label{sec:back}

% While \dalton can solve ODEs of arbitrary order, for ease of presentation we focus on first-order systems.  The extension to higher-order ODEs is provided in Appendix~\appref{sec:higher}.
\dalton is designed to solve arbirary-order multi-variable ODE systems which satisfy an initial value problem (IVP).  For ease of presentation we focus on first-order systems.  The extension to higher-order problems is described in Appendix~\appref{sec:higher}. 
For a multi-variable function $\xx(t) = (x_1(t), \ldots, x_d(t))$, a first-order ODE-IVP is of the form
\begin{equation}\label{eq:mode}
    \dot \xx(t) = \ff(\xx(t), t), \qquad \xx(0) = \vv, \qquad t \in [0, T],
\end{equation}
where $\dot \xx(t) = \dv{t} \xx(t)$ is the first derivative of $\xx(t)$, $\ff(\xx(t), t) = (f_1(\xx(t), t), \ldots, f_d(\xx(t), t))$ is a nonlinear function, and $\vv = \big(v_1, \ldots, v_d \big)$ is the initial value of $x(t)$ at time $t = 0$.

Unlike deterministic solvers, \dalton employs a probabilistic approach to solving~\eqref{eq:mode} based on a well-established paradigm of Bayesian nonlinear filtering~\citep{tronarp.etal18,schober.etal19,schmidt21,tronarp22}.  This approach consists of putting a Gaussian Markov process on $\xx(t)$ and its first $q-1$ derivatives,
\begin{equation}\label{eq:xprior}
  \textstyle \XX(t) = (\xx(t), \dot \xx(t), \ldots, \dv[q-1]{t}\xx(t)),
\end{equation}
% embedding $(\xx(t), \dot \xx(t))$ into a Gaussian Markov process $\XX(t)$,  
% consists of putting a Gaussian Markov process prior on $\XX(t) = (x_1(t), \dot x_1(t), \ldots x_d(t), \dot x_d(t))$,
and updating $\XX(t)$ with information from the ODE-IVP~\eqref{eq:mode} at time points $t = t_0, \ldots, t_N$, where $t_n = n \cdot \dt$ and $\dt = T/N$.  Specifically, let $\XX_n = \XX(t_n)$ and consider the general indexing notation $\XX_{m:n} = (\XX_m, \ldots, \XX_n)$.  If $\xx(t)$ is the solution to~\eqref{eq:mode}, we would have $\ZZ_n = \dot \xx_n - \ff(\xx_n, t_n) = \bz$. Based on this observation, \cite{tronarp.etal18} consider a state-space model on $\XX_n$ and $\ZZ_n$ of the form
\begin{equation}\label{eq:bnf}
    \begin{aligned}
        \XX_{n+1} \mid \XX_n & \sim \N(\QQ \XX_n, \RR) \\
        \ZZ_n & \ind \N(\dot \xx_n - \ff(\xx_n, t_n), \VV),
    \end{aligned}
\end{equation}
where
% $\XX_0 = (v_1, \ff_1(v_1, t_0), \ldots, v_d, \ff_d(v_d, t_0))$,
$\QQ = \QQ(\dt)$ and $\RR = \RR(\dt)$ are determined by the Gaussian Markov process prior on $\XX(t)$.
% depending on the tuning parameter $\eet$.
The specific Gaussian Markov process prior used in this work is described in Appendix~\appref{sec:prior}. The stochastic ODE solution is then given by the posterior distribution $p(\XX_{0:N} \mid \ZZ_{0:N} = \bz)$ resulting from model~\eqref{eq:bnf}. 

As $N \to \infty$ and $\VV \to \bz$, the mode of the posterior distribution $p(\XX_{0:N} \mid \ZZ_{0:N} = \bz)$ gets arbitrarily close to the true ODE solution~\citep{tronarp.etal21}. However, this posterior distribution cannot be sampled from directly unless $\ff(\xx(t), t)$ is a linear function. Alternatives methods include Markov chain Monte Carlo (MCMC) sampling~\citep{yang.etal21} and particle filtering~\citep{tronarp.etal18}. A less accurate but ostensibly much faster approach is to linearize~\eqref{eq:bnf}, resulting in the working model
\begin{equation}\label{eq:lin}
    \begin{aligned}
        \XX_{n+1} \mid \XX_n & \sim \N(\QQ \XX_n, \RR) \\
        \ZZ_n & \ind \N(\dot \xx_n + \BB_n \xx_n + \aa_n, \VV_n),
    \end{aligned}
\end{equation}
where $\VV_n$ is a tuning parameter used to capture the discrepancy between~\eqref{eq:lin} and~\eqref{eq:bnf} with $\VV = \bz$. The benefit of the linearized model~\eqref{eq:lin} is that it gives an approximation to (i) the posterior distribution $p(\XX_{0:N} \mid \ZZ_{0:N} = \bz)$, and (ii) the marginal likelihood $p(\ZZ_{0:N} = \bz)$ 
% -- and , which we will need in Section~\ref{sec:meth} --
having linear complexity $\bO(N)$ in the number of time discretization points, using standard Kalman filtering and smoothing techniques~\citep{tronarp.etal18,schober.etal19}. Many linearization approaches can be found in~\citep{tronarp.etal18, kramer21} and different choices for $\VV_n$ are discussed in~\citep{schober.etal19, kersting20}.  Each determines the linearization parameters $(\aa_n, \BB_n, \VV_n)$ at time $t = t_n$ 
% in terms of the mean and variance $\mmu_{n|n-1}$ and $\SSi_{n|n-1}$ of the Gaussian predictive distribution
% at time $t = t_n$
in terms of the mean and variance of the Gaussian predictive distribution
% $p_\L(\xx_n \mid \ZZ_{0:n-1} = \bz)$ induced by the linearized working model~\eqref{eq:lin} up to time $t = t_{n-1}$, which is determined by the predicted mean and variance
\begin{equation}\label{eq:linpred}
  % \begin{aligned}
  %   \mmu_{n|n-1} & = \E_\L[\xx_n \mid \ZZ_{0:n-1} = \bz], \\ \SSi_{n|n-1} & = \var_\L(\xx_n \mid \ZZ_{0:n-1} = \bz).
  % \end{aligned}
  p_\L(\xx_n \mid \ZZ_{0:n-1} = \bz) % \qquad \iff \qquad \xx_n \sim \N(\mmu_{n|n-1}, \SSi_{n|n-1})
\end{equation}
induced by the linearized working model~\eqref{eq:lin} up to time $t = t_{n-1}$.
Perhaps the simplest of these linearization schemes uses a zeroth-order Taylor approximation to $\ff(\xx_n, t_n)$~\citep{schober.etal19}, with
\begin{equation}\label{eq:zero}
    \begin{aligned}
        \aa_n & =  \ff(\mmu_{n|n-1}, t_n), & \BB_n & = \bz, & \VV_n & = \bz,
    \end{aligned}
  \end{equation}
  where $\mmu_{n|n-1} = \E_\L[\xx_n \mid \ZZ_{0:n-1} = \bz]$.
% can be computed efficiently using a Kalman filter.
% is the predicted mean obtained sequentially from a Kalman filter applied to~\eqref{eq:lin}.
It has been shown that as $N \to \infty$, the mean and variance of the forward distribution $p_\L(\xx_n \mid \ZZ_{0:n} = \bz)$ induced by the linearization~\eqref{eq:zero} converge, respectively, to the true ODE solution and zero~\citep{kersting20}.
% has $\VV_n = \bz$ and uses a zeroth order Taylor approximation for the nonlinear ODE function,
% \begin{equation}\label{eq:zero}
%     \begin{aligned}
%         \ff(\xx_n, t_n) \approx \ff(\hat \xx_n, t_n),
%     \end{aligned}
% \end{equation}
% where $\hat \xx_n = \E[\xx_n \mid \ZZ_{0:n-1}]$ is the predicted mean obtained sequentially from a Kalman filter applied to~\eqref{eq:lin}, i.e., with $\aa_n = -\ff(\hat \xx_n, t_n)$ and $\BB_n = \bz$. In particular, it has been proven by~\cite{kersting20} that as $N \to \infty$ the mean and variance computed from a Kalman filter using the linearization of~\eqref{eq:zero} converges to the true ODE solution and zero respectively.

\section{Methodology}\label{sec:meth}

The parameter-dependent extension of the ODE-IVP~\eqref{eq:mode} 
% for a multi-variable function $\xx(t) = \big(x_1(t), \ldots, x_d(t)\big)$
is of the form
\begin{equation}\label{eq:pode}
    \dot \xx(t) = \ff_\tth(\xx(t), t), \qquad \xx(0) = \vv_\tth, \qquad t \in [0, T].
\end{equation}
% For the extension to higher-order ODE systems, please see Appendix~\appref{sec:higher}
The learning problem consists of estimating the unknown parameters of the model $\tth$ which determine $\xx(t)$ in~\eqref{eq:pode} from noisy observations $\YY_{0:M} = (\YY_0, \ldots, \YY_M)$, recorded at times $t = t'_0, \ldots, t'_M$ under the measurement model
\begin{equation}\label{eq:meas}
    \YY_i \ind p(\YY_i \mid \xx(t'_i), \pph),
\end{equation}
with possibly unknown model parameters $\pph$. In terms of the ODE solver discretization time points $t = t_0, \ldots, t_N$, $N \ge M$, consider the mapping $n(\cdot)$ such that $t_{n(i)} = t'_i$, and the inverse mapping $i(n) = \max\{i: n(i) \le n\}$.  \dalton then augments the Bayesian filtering model~\eqref{eq:bnf} to account for noisy observations from~\eqref{eq:meas} via
% where $\DD_i$ is a known matrix of zeros and ones used to select the a coefficient matrix which selects the partially observed components of $\XX(t)$. $\XX_{0:N}$ and $\ZZ_{0:N}$ correspond to time points $t = t_0, \ldots, t_N$, with $N \ge M$ and a mapping $n(\cdot)$ such that $t_{n(i)} = t'_i$.
% Directly accounting for the observations, \dalton assumes the model
\begin{equation}\label{eq:dalton}
  \begin{aligned}
    \XX_{n+1} \mid \XX_n & \sim \N(\QQ_\eet \XX_n, \RR_\eet) \\
    \ZZ_n & \ind \N(\dot \xx_n - \ff_\tth(\xx_n, t_n), \VV) \\
    \YY_i & \ind p(\YY_i \mid \xx_{n(i)}, \pph),
  \end{aligned}
\end{equation}
where $\eet$ are tuning parameters of the Gaussian Markov process prior. 
The likelihood function induced by the probabilistic solver corresponding to~\eqref{eq:dalton} for all parameters $\TTh = (\tth, \pph, \eet)$ is given by
% , which are also to be estimated from the data.  to calibrate the Gaussian Markov process prior to the data as well. There are other potential methods for calibration~\citep[e.g.][]{schober.etal19}, but we follow the method as proposed by~\citet{chkrebtii.etal16}. Ultimately, our goal is to compute the likelihood-based parameter inference given by
\begin{equation}\label{eq:likepar}
    \mathcal{L}(\TTh \mid \YY_{0:M}) = p(\YY_{0:M} \mid \ZZ_{0:N} = \bz, \TTh).
\end{equation}

\subsection{Gaussian Measurement Model}\label{sec:gauss}

First suppose the observations of~\eqref{eq:meas} consist of Gaussian noise,
\begin{equation}\label{eq:gmeas}
    \YY_i \ind \N(\DD_i^\pph \xx_{n(i)}, \OOm_i^\pph),
\end{equation}
where $\DD_i^\pph$ and $\OOm_i^\pph$ are (possibly $\pph$-dependent) coefficient and variance matrices.  To compute the likelihood~\eqref{eq:likepar}, we begin with the identity
\begin{equation}\label{eq:condp}
    p(\YY_{0:M} \mid \ZZ_{0:N} = \bz) = \frac{p(\YY_{0:M}, \ZZ_{0:N} = \bz)}{p(\ZZ_{0:N} = \bz)},
\end{equation}
where we have omitted the dependence on $\TTh$ for notational brevity.  The denominator $p(\ZZ_{0:N} = \bz)$ on the right-hand side can be approximated using the Kalman recursions applied to the linearized state-space model~\eqref{eq:lin}.  In fact, the same can be done for the numerator $p(\YY_{0:M}, \ZZ_{0:N} = \bz)$.  That is, one begins by linearizing the measurement model~\eqref{eq:dalton} via
\begin{equation}\label{eq:pling}
  \begin{aligned}
    \XX_{n+1} \mid \XX_n & \sim \N(\QQ_\eet \XX_n, \RR_\eet) \\
    \ZZ_n & \ind \N(\dot \xx_n + \BB_n \xx_n + \aa_n, \VV_n) \\
    \YY_i & \ind \N(\DD_i^\pph \xx_{n(i)}, \OOm_i^\pph),
  \end{aligned}
\end{equation}
and using any of the linearization approaches for the data-free setting in Section~\ref{sec:back}, but applied to the mean and variance of the Gaussian predictive distribution induced by~\eqref{eq:pling},
\begin{equation}\label{eq:plindist}
  p_\L(\xx_n \mid \YY_{0:i(n-1)}, \ZZ_{0:n-1} = \bz),
\end{equation}
% where $i(n) = \max\{i: n(i) \le n\}$,
which conditions both on information from the ODE and the measurements up to time $t = t_{n-1}$.
The complete \dalton algorithm for estimating $p(\YY_{0:M} \mid \ZZ_{0:N} = \bz, \TTh)$ is provided in Algorithm~\appref{alg:dalton} of Appendix~\appref{sec:alg}, in terms of the standard Kalman filtering and smoothing recursions detailed in Appendix~\appref{sec:kalmanfun}.

\subsection{Non-Gaussian Measurement Model }\label{subsec:ngdalton}

Let us now turn to the general measurement model~\eqref{eq:meas}.  Let $\xobs_{n(i)}$ and $\xun_{n(i)}$ denote observed and unobserved components of $\xx(t)$ at time $t = t_i'$, such that \mbox{$\pdv{\xun_{n(i)}} p(\YY_i \mid \xx_{n(i)}, \pph) = \bz$} and
%\begin{equation}\label{eq:nongmeas}
%  \YY_i \ind \exp\{-g(\YY_i \mid \DD_i \XX_{n(i)}, \pph)\},
%\end{equation}
%where $\DD_i$ is a mask matrix of size $s_i \times q_0$, where $q_0 = \sum_{k=1}^d q_k$ and $s_i \le q_0$.  In other words, $\DD_i$ is a subset of the rows of the $q_0 \times q_0$ identity matrix, which is used to select the partially observed components of $\XX(t)$.  
%\textbf{Note:} I think the observation model should be
\begin{equation}\label{eq:nongmeas}
    p(\YY_i \mid \xx_{n(i)}, \pph) =  \exp\{-g_i(\YY_i \mid \xobs_{n(i)}, \pph)\}.
\end{equation}
% where $\xobs_{n(i)}$ is the subset of $\XX_{n(i)}$ corresponding to the partially observed components of $\xx(t)$, which may depend on the time point $t = t_{n(i)}$.
In order to compute the likelihood~\eqref{eq:likepar}, again omitting the dependence on $\TTh$ we consider a different identity,
\begin{equation}\label{eq:nongdec}
  \begin{aligned}
    & \phantom{\;=\;} p(\YY_{0:M} \mid \ZZ_{0:N} = \bz) =  \frac{p(\YY_{0:M}, \XX_{0:N} \mid \ZZ_{0:N} = \bz)}{p(\XX_{0:N} \mid \YY_{0:M}, \ZZ_{0:N} = \bz)} \\
    & = \frac{p(\XX_{0:N} \mid \ZZ_{0:N} = \bz) \times \prod_{i=0}^M \exp\{- g_i(\YY_i \mid \xobs_{n(i)})\}}{p(\XX_{0:N} \mid \YY_{0:M}, \ZZ_{0:N} = \bz)},
  \end{aligned}
\end{equation}
where the identity holds for any value of $\XX_{0:N}$.  In the numerator of~\eqref{eq:nongdec}, $p(\XX_{0:N} \mid \ZZ_{0:N} = \bz)$ can be approximated using the Kalman smoothing algorithm applied to the data-free linearized model~\eqref{eq:lin}, whereas the product term is obtained via straightforward calculation of~\eqref{eq:nongmeas}.  As for the denominator of~\eqref{eq:nongdec}, we propose to approximate it by a multivariate normal distribution as follows.  First, we note that 
\begin{equation}\label{eq:logpxy}
    \begin{aligned}
        &\log p(\XX_{0:N} \mid \YY_{0:M}, \ZZ_{0:N}=\bz) \\
        &= \log p(\XX_{0:N} \mid \ZZ_{0:N}=\bz) - \sum_{i=0}^M h_i(\xobs_{n(i)}) + c_1,
        % g_i(\YY_i \mid \xobs_{n(i)}) + c_1,
\end{aligned}
\end{equation}
where $h_i(\xobs) = g_i^\pph(\YY_i \mid \xobs)$ and $c_1$ is constant with respect to $\XX_{0:N}$.  Next, we take a second-order Taylor expansion of $h_i(\xobs_{n(i)})$ about a value $\xobs = \hxobs_{n(i)}$ to be specified momentarily.  After simplification this gives
\begin{equation}\label{eq:taylorapp}
    h_i(\xobs) \approx - \frac 1 2 (\xobs - \hat \YY_i)'\nabla^2 h_i(\hxobs_{n(i)}) (\xobs - \hat \YY_i) + c_2,
\end{equation}
where $\nabla h_i(\xobs)$ and $\nabla^2 h_i(\xobs)$ are the gradient and Hessian of $h_i(\xobs)$, $\hat \YY_i = \hxobs_{n(i)} - \nabla^2 h_i(\hxobs_{n(i)})^{-1}\nabla h_i(\hxobs_{n(i)})$, and $c_2$ is constant with respect to $\xobs$.  Substituting the quadratic approximation~\eqref{eq:taylorapp} into~\eqref{eq:logpxy}, we obtain an estimate of $p(\XX_{0:N} \mid \YY_{0:M}, \ZZ_{0:N} = \bz)$ equivalent to that of a model with Gaussian observations,
\begin{equation}\label{eq:pgauss}
  \begin{aligned}
    \XX_{n+1} \mid \XX_n & \sim \N(\QQ_\eet \XX_n, \RR_\eet) \\
    \ZZ_n & \ind \N(\dot \xx_n - \ff_\tth(\xx_n, t_n), \VV) \\
    \hat \YY_i & \ind \N(\xobs_{n(i)}, [\nabla^2 h_i(\hxobs_{n(i)})]^{-1}).
  \end{aligned}
\end{equation}
We may now linearize~\eqref{eq:pgauss} exactly as in Gaussian setting of Section~\ref{sec:gauss} to obtain an estimate
\begin{equation}\label{eq:pgausslin}
  p_\L(\XX_{0:N} \mid \hat \YY_{0:M}, \ZZ_{0:N} = \bz),
\end{equation}
which may be computed efficiently using the Kalman smoother.  
% and using the data-free linearization of $p(\XX_{0:N} \mid \ZZ_{0:N} = \bz, \tth, \eet)$, we may estimate $p(\XX_{0:N} \mid \YY_{0:M}, \ZZ_{0:N} = \bz, \TTh)$ as the multivariate normal arising from the working model
% \begin{equation}\label{eq:plin}
%   \begin{aligned}
%     \XX_{n+1} \mid \XX_n & \sim \N(\QQ_\eet \XX_n, \RR_\eet) \\
%     \ZZ_n & \ind \N(\dot \xx_n \BB_n \xx_n + \aa_n, \VV_n) \\
%     \hat \YY_i & \ind \N(\xx_{n(i)}, [\nabla^2 h_i(\hat \xx_{n(i)})]^{-1}),
%   \end{aligned}
% \end{equation}
% where the data-free linearization coefficients $\aa_{n(i)}$, $\BB_{n(i)}$, and $\VV_{n(i)}$ are computed exactly as in~\eqref{eq:lin}.  Calculation of $p(\XX_{0:N} \mid \YY_{0:M}, \ZZ_{0:N} = \bz, \TTh)$ the combines the Kalman forward pass of Algorithm~\eqref{alg:dalton} with the backward pass of a standard Kalman smoother.
To complete the algorithm, there remains the choice $\XX_{0:N}$ to plug into~\eqref{eq:nongdec}, and the choice of $\hxobs_{n(i)}$ about which to perform the Taylor expansion~\eqref{eq:taylorapp}.  For the latter, we use
% \begin{equation}
  $\hxobs_{n(i)} = \E_\L[\xobs_{n(i)} \mid \hat \YY_{0:i-1}, \ZZ_{0:n(i)-1} = \bz]$, 
% \end{equation}
the predicted mean of the linearized model obtained from~\eqref{eq:pgauss}.  For the former, we use
% \begin{equation}
  $\XX_{0:N} = \E_\L[\XX_{0:N} \mid \hat \YY_{0:M}, \ZZ_{0:N} = \bz]$,
% \end{equation}
the mean of~\eqref{eq:pgausslin},
% conditional mean of $\XX_{0:N}$ given all the observed data in the working model~\eqref{eq:plin},
which can also be computed efficiently using the Kalman recursions.
% is readily obtained from the Kalman filtering and smoothing recursions.
Full details of the \dalton algorithm for non-Gaussian measurements are provided in Algorithm~\appref{alg:daltonng} of Appendix~\appref{sec:alg}.

%\subsection{Heuristic Justification}\label{sec:hjust}
%
%We give a brief heuristic justification for the convergence of the \dalton approximation with arbitrary noise~\eqref{eq:nongdec}-\eqref{eq:plin} to the true likelihood as $N \to \infty$ with $\VV_n = \bz$. First, it has been proved by~\cite{kersting20} that $p(\XX_{0:N} \mid \ZZ_{0:N}, \tth, \eet)$ concentrates on the true ODE solution under the linearization~\eqref{eq:zero}.  We assume that the same holds true for $p(\XX_{0:N} \mid \YY_{0:M}, \ZZ_{0:N} = \bz, \TTh)$ under the extended linearization of~\eqref{eq:plin}.  Short of a formal proof, an informal justification is that $\YY_{0:M}$ adds only a finite amount of information to the posterior on $\XX_{0:N}$, whereas $\ZZ_{0:N}$ adds an infinite amount as $N \to \infty$, thus ultimately overwhelming the information provided by the observed data. Since these two posteriors converge to the same value, we expect them to cancel out in~\eqref{eq:nongdec}, leaving just $\prod_{i=0}^M \exp\{-g_i(\YY_i \mid \xx_{n(i)}, \pph)\}$ in the numerator of~\eqref{eq:nongdec} with $\XX_{0:N}$ the true ODE solution, which is precisely the form the true ODE likelihood.

\section{Related Work}\label{sec:related}

The Bayesian filtering paradigm~\eqref{eq:bnf} was formulated by~\cite{tronarp.etal18}.  Speed and stability improvements on the linearizations presented therein were developed in~\cite{kramer20,kramer21}.  Various convergence properties may be found in~\cite{chkrebtii.etal16,schober.etal19,kersting20}.

\textbf{Gradient matching} with Gaussian processes~\citep{calderhead.etal09} is an early precursor to the Bayesian filtering paradigm~\eqref{eq:bnf}.  Similar in spirit to deterministic collocation solvers~\citep[e.g.,][]{ramsay.etal07}, a conceptual limitation of this approach is the lack of a generative model on both the Gaussian process $\XX(t)$ and the observed data $\YY_{0:M}$~\citep{barber.wang14}, leading to parameter identifiability issues~\citep{macdonald.etal15,wenk.etal19}.  Moreover, many gradient-matching approaches involve computationally-intensive methods such as MCMC~\citep{barber.wang14,lazarus.etal18} and approximate Bayesian computation (ABC)~\citep{ghosh.etal17}, though computationally efficient variational approximations have been developed as well~\citep[e.g.,][]{dondelinger.etal13,gorbach.etal17,wenk.etal19}.

\textbf{Active uncertainty calibration}~\citep{kersting.hennig16} formulates the ODE solver as a Gaussian process filtering problem.  Not only does this approach allow to naturally extend many deterministic ODE solvers~\citep[e.g.,][]{schober.etal14,teymur.etal16,schober.etal19,bosch22}, it also admits analytic calculations which scale linearly in the number of discretization points $N$ when a Markov Gaussian process prior is employed.  Active uncertainty calibration is operationally equivalent to linearization of the Bayesian filtering model~\eqref{eq:bnf}, though the latter has the conceptual benefit of separating the desired solution posterior $p(\XX_{0:N} \mid \ZZ_{0:N})$ from the linearization~\eqref{eq:lin} used to approximate it.

% - Gradient matching with GPs: combining GP model for solution + data with ODE restrictions on solution \cite{calderhead.etal09, dondelinger.etal13, barber.wang14, gorbach.etal17, ghosh.etal17, lazarus.etal18, wenk.etal19}.  Not clear how to make these two models compatible, and parameter identifiability issue \cite{macdonald.etal15}.

% - Active uncertainty calibration: The next step towards resolving the incompatibility issue \cite{schober.etal14, kersting.henning16, teymur.etal16, chkrebtii.etal16, schober.etal19}.  Utilizes a Gaussian updating scheme to provide analytical calculations.  However, combines both the model and linearization of \cite{tronarp.etal18}.

% - Within Bayesian filtering paradigm, highlight the following approaches which condition directly on the data.

%Various probabilistic ODE solvers have been presented in the works of e.g.,~\cite{calderhead.etal09,dondelinger.etal13,barber.wang14,schober.etal14,chkrebtii.etal16,kersting.hennig16,teymur.etal16,conrad.etal17,ghosh.etal17,gorbach.etal17,lazarus.etal18,tronarp.etal18,schober.etal19,wenk.etal19,kramer20,kersting20a,yang.etal21,wenger21,kramer21,bosch21,tronarp22,bosch22}.  The Bayesian filtering paradigm~\eqref{eq:bnf} at the heart of many of these solvers was formulated in~\cite{tronarp.etal18}, alternatively viewed as a predictor-corrector method in~\cite{chkrebtii.etal16,schober.etal19}.  Various convergence properties of the corresponding probabilistic solver are derived in~\cite{chkrebtii.etal16,schober.etal19,kersting20}.

Within the Bayesian filtering framework, we highlight the following approaches in which the ODE solver directly conditions on the observed data:

\textbf{\cite{chkrebtii.etal16}} uses active uncertainty calibration to stochastically estimate the optimal linearization parameter $\aa_n$ in the sense of~\cite{kersting.hennig16,tronarp.etal18}. However, the nondeterministic $\aa_n$ cannot be analytically integrated out, such that inference for $\TTh$ is conducted by sampling from the joint posterior $p(\TTh, \XX_{0:N} \mid \YY_{0:M}, \ZZ_{0:N} = \bz)$ uses MCMC techniques.
% , which requires at least an order of magnitude more evaluations of the probabilistic ODE solver than the \dalton-based parameter learning method presented in Section~\ref{sec:ex}.
%slower than the approximate Bayesian a similar approach in the predictor-corrector formulation. Both methods require Markov chain Monte Carlo (MCMC) techniques to sample from the desired posterior, which requires at least an order of magnitude times more evaluations of the probabilistic ODE solver than the approximate Bayesian method based on \dalton we present in Section~\ref{sec:ex}.

\textbf{MAGI}~\citep{yang.etal21} uses a closed-form expression for the posterior distribution $p(\TTh, \XX_{0:N} \mid \YY_{0:M}, \ZZ_{0:N} = \bz)$ resulting from the Bayesian filtering model~\eqref{eq:dalton} with ``perfect'' ODE information $\VV = \bz$. 
% , by carefully defining the conditioning on a set of measure zero to avoid the Borel-Kolmogrov paradox.
Again this requires sampling from the joint posterior using MCMC techniques.

%estimates $\TTh$ in a Bayesian context from the exact posterior $p(\TTh \mid \YY_{0:M}, \ZZ_{0:N} = \bz)$ resulting from~\eqref{eq:dalton} with $\VV_n = \bz$. One limitation to this method is that the dimension of the underlying matrix scales with the time discretization points resulting in possible memory issues.

\textbf{\cite{schmidt21}} uses a data-driven extended Kalman filter similar to ours, but focuses on estimating the ODE solution itself rather than the model parameters.

\textbf{Fenrir}~\citep{tronarp22} extends an approach developed in~\cite{kersting20a}.  It begins by using the data-free linearization of~\eqref{eq:lin} to estimate $p(\XX_{0:N} \mid \ZZ_{0:N} = \bz, \tth, \eet)$. This posterior can be sampled from using a (non-homogeneous) Markov chain going backwards in time,
\begin{equation}\label{eq:fenrirback}
    \begin{aligned}
        \XX_N & \sim \N(\bb_N, \CC_N) \\
        \XX_n \mid \XX_{n+1} & \sim \N(\AA_n \XX_n + \bb_n, \CC_n),
    \end{aligned}
\end{equation}
where the coefficients $\AA_{0:N-1}$, $\bb_{0:N}$, and $\CC_{0:N}$ are derived using the Kalman recursions~\citep{tronarp22}. Next, \fenrir assumes that Gaussian observations are added to the model, for which the marginal likelihood 
\begin{equation}\label{eq:fenrirlike}
    \begin{aligned}
        &p(\YY_{0:M} \mid \ZZ_{0:N} = \bz, \TTh)\\
        &=\int p(\YY_{0:N} \mid \XX_{0:N}, \TTh) p(\XX_{0:N} \mid \ZZ_{0:N}= \bz, \TTh) \ud \XX_{0:N}
    \end{aligned}
\end{equation}
is computed using a Kalman filter on the backward pass of~\eqref{eq:fenrirback}. \fenrir is fast, accurate, and compares favorably to both the fast gradient-based solver of~\cite{wenk.etal19} and various
% gradient matching based probabilistic solver, \textbf{ODIN}~\citep{wenk.etal19} and
deterministic ODE solvers~\citep{tronarp22}. The key difference between \dalton and \fenrir is that the latter linearizes \emph{before} adding the observations to the model (linearize-then-observe), whereas the former does so \emph{after} (observe-then-linearize).  Moreover, \dalton can be applied to non-Gaussian errors whereas \fenrir cannot.

\paragraph{Computational Scaling}
\dalton scales linearly in both $N$, the number of time discretization points, and $d$, the dimension of the ODE solution $\xx(t)$.  The former is due to the Markov structure of the prior on $\XX(t)$, whereas the latter is achieved by using the blockwise linearization of~\cite{kramer21} (details in Appendix~\ref{sec:first}). %scales linearly in both time discretization points $N$ and ODE variables $d$ by utilizing the block diagonal structure of the matrices~\citep{kramer21} in its prior and linearization (full details provided in Appendix~\ref{sec:prior},~\ref{sec:first}).

For the Gaussian measurement model~\eqref{eq:gmeas}, the operation count of 
% In terms of timing comparisons,
\dalton and \fenrir is roughly the same, with \fenrir doing one forward and one backward pass through the $N$ discretization points, % on $t_{0:N}$,
and \dalton doing one forward pass
% for the numerator and one for the denominator of~\eqref{eq:condp}. %
for each of $p(\ZZ_{0:N} = \bz \mid \TTh)$ and $p(\YY_{0:M}, \ZZ_{0:N} = \bz \mid \TTh)$ in~\eqref{eq:condp}. For non-Gaussian measurements, \dalton requires roughly twice as many operations as in the Gaussian case, % The non-Gaussian \dalton of Algorithm~\ref{alg:daltonng} is about twice as slow as the Gaussian \dalton of Algorithm~\ref{alg:dalton},
since the former requires one forward and backward pass for each of $p(\XX_{0:N} \mid \ZZ_{0:N} = \bz, \tth, \eet)$ and $p(\XX_{0:N} \mid \YY_{0:M}, \ZZ_{0:N} = \bz, \TTh)$ in~\eqref{eq:nongdec}.

% Both \magi and \chk are an order of magnitude slower ($\approx$ 10 times) because they require MCMC techniques to sample the posterior which require more evaluations of the probabilistic ODE solver than the Laplace approximation.

\section{Examples}\label{sec:ex}

We now evaluate the performance of \dalton in several numerical examples of parameter learning.  We proceed with a Bayesian approach by postulating a prior distribution $\pi(\TTh)$ on the full set of parameters $\TTh = (\tth, \pph, \eet)$, which combined with~\eqref{eq:likepar} gives the \dalton posterior
\begin{equation}\label{eq:dalpost}
  \begin{aligned}
    p(\TTh \mid \YY_{0:M}) & \propto \pi(\TTh) \times \mathcal{L}(\TTh \mid \YY_{0:M}) \\
                           & = \pi(\TTh) \times p(\YY_{0:M} \mid \ZZ_{0:N} = \bz, \TTh).
  \end{aligned}
\end{equation}
We take the measurement model parameter $\pph$ to be known. Parameter inference is then accomplished by way of a Laplace approximation~\citep[e.g.,][]{gelman.etal13}, for which we have
\begin{equation}\label{eq:bna}
    \TTh \mid \YY_{0:M} \approx \N(\hat \TTh, \hat \VV_{\TTh}),
\end{equation}
where $\hat \TTh = \argmax_{\TTh} \log p(\TTh \mid \YY_{0:M})$ and $\hat \VV_{\TTh} = -\big[\pdv{}{\TTh}{\TTh'} \log p(\hat \TTh \mid \YY_{0:M})\big]^{-1}$. The Laplace approximation is a popular tool for Bayesian machine learning applications~\citep{mackay92, gianniotis19}, typically requiring at least an order of magnitude fewer evaluations of $p(\TTh \mid \YY_{0:M})$ than full Bayesian inference via MCMC.  Our \python implementation of \dalton\appfootnote uses the \jax library~\citep{jax2018github} for automatic differentiation and just-in-time (JIT) compilation to obtain a very fast implementation of the Laplace approximation~\eqref{eq:bna}. % (full details provided in Appendix~\ref{sec:parpost}).

Data for the examples are simulated using a high-accuracy deterministic solver for the ODE-IVP~\eqref{eq:pode}, namely, the Dormand-Prince 8/7 solver consisting of an 8th order Runge-Kutta method with 7th order step size adaptation (hereafter \rk)~\citep{prince.dormand81}.
% solver with Dormand-Prince step size adaptation~\citep{dormand.prince80}
% as implemented in the \diffrax library~\citep{kidger21}.
We compare the parameter inference results of \dalton to three competing probabilistic ODE likelihood methods: (i) a Laplace approximation with the \fenrir likelihood, and full Bayesian inference via MCMC for the (ii) \magi and (iii)~\cite{chkrebtii.etal16} methods (hereafter \chk).  We also compare to (iv) a Laplace approximation using the deterministic \rk solver, which we assume to provide the true ODE solution.  We use our own implementations of \dalton, \fenrir, and \chk, the \magi implementation provided by~\cite{wong2022magi}\footnote{https://github.com/wongswk/magi}, and the \rk implementation provided~\cite{kidger21}\footnote{https://github.com/patrick-kidger/diffrax}.  The \chk and \magi algorithms are run for 100,000 and 10,000 iterations each (post burn-in), which produced effective samples sizes
% across all experiments
ranging between 100-5000.  Further implementation details for the parameter learning methods are provided in Appendix~\ref{sec:parpost}.
%by th described in.  We also compare to a Laplace approximation using the high-accuracy deterministic \rk ODE solver provided by \diffrax. We shall assume that the output of \rk is the true ODE solution.

\subsection{FitzHugh-Nagumo Model}\label{sec:fitz}

\begin{figure}[!htb]
     \includegraphics[width=\linewidth]{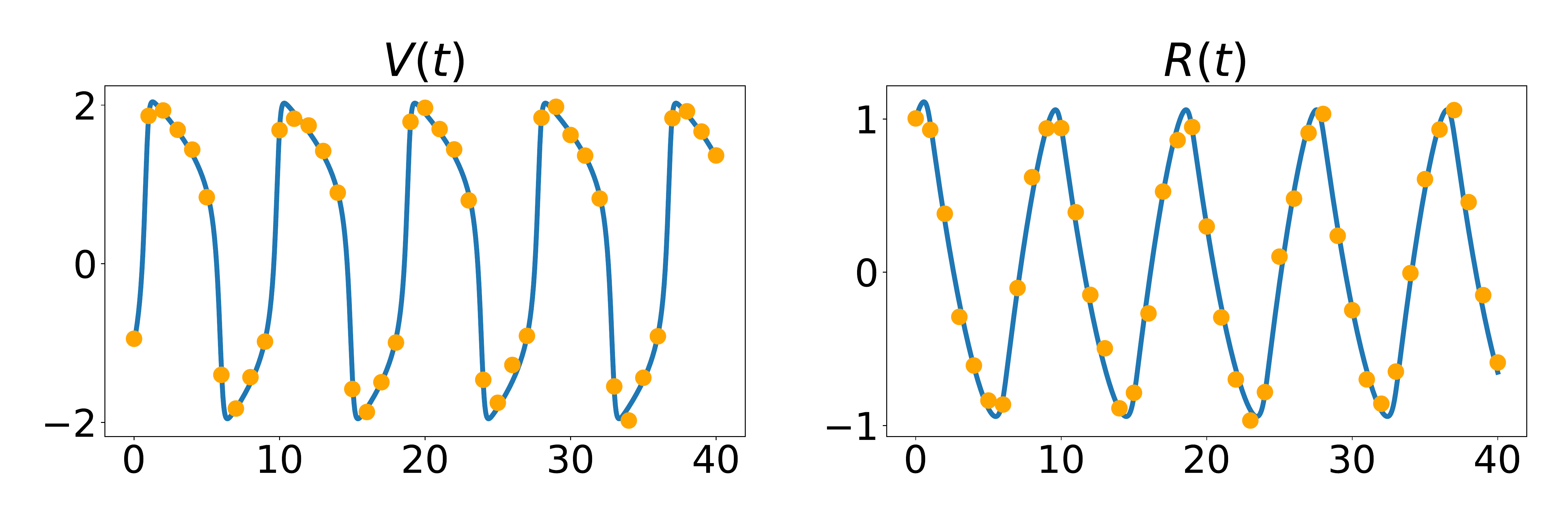}
    \caption{
        ODE and noisy observations for the FN model.
    }
    \label{fig:fitzdata}
\end{figure}

The FitzHugh-Nagumo (FN) model~\citep{fitzhugh61,nagumo.etal62} is a two-state ODE on $\xx(t) = (V(t), R(t))$,
in which $V(t)$ describes the evolution of the neuronal membrane voltage and $R(t)$ describes the activation and deactivation of neuronal channels. The FN ODE is given by
\begin{equation} \label{eq:fitz}
    \begin{aligned}
        \dot V(t) &= c\Big(V(t) - \frac{V(t)^3}{3} + R(t)\Big), \\
        \dot R(t) &= -\frac{V(t) - a + bR(t)}{c}.
    \end{aligned}
\end{equation}
The model parameters are $\tth = (a, b, c, V(0), R(0))$ with $a, b, c > 0$. These are to be learned from the measurement model
\begin{equation}\label{eq:fitznoise}
    \YY_i \ind \N(\xx(t_i), \phi^2 \cdot \Id_{2\times 2}),
\end{equation}
where $t_i = i$ with $i=0, 1, \ldots 40$ and $\phi^2 = 0.005$. The ODE and noisy observations are displayed in Figure~\ref{fig:fitzdata}.

\begin{figure}[!htb]
 \includegraphics[width=\linewidth]{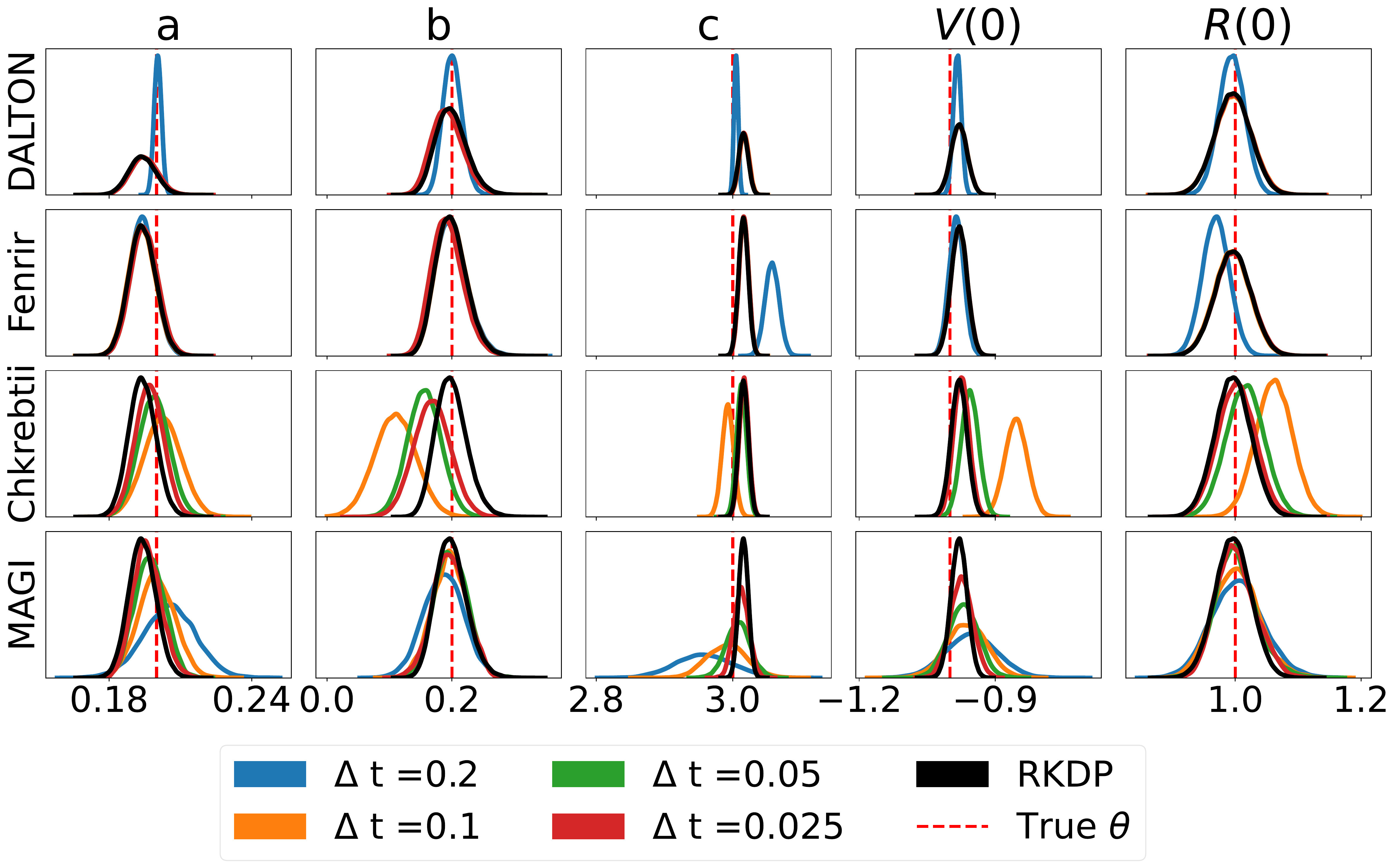}
 \caption{
     Parameter posteriors for the FN model using various ODE solvers.
 }
 \label{fig:fitzpost}
\end{figure}

Figure~\ref{fig:fitzpost} displays the 
\dalton, \fenrir, \chk, and \magi posteriors 
% Laplace posteriors for \dalton and \fenrir and the MCMC samples of \chk and \magi
for data simulated with true parameter values $\tth = (0.2, 0.2, 3, -1, 1)$, at different values of the step size $\dt = T/N$. Also included for comparison is the Laplace posterior for the true likelihood obtained with \rk. 
% are Laplace posteriors with \fenrir for the same step sizes, and for the true likelihood obtained with \rk.  
% The black posteriors is the ground truth produced by a highly accurate deterministic solver and the red dotted lines are the true parameters.
At smaller step sizes, all posterior distributions are virtually identical.  At larger step sizes, \magi and \chk are further from the \rk posterior than either \dalton or \fenrir, which \chk failing to converge at the largest step size $\dt = 0.2$.  For $\dt = 0.2$, \dalton and \fenrir posteriors are similar, with the notable exception of the posterior for $c$, where \dalton covers the true parameter value but \fenrir does not.  This is because $c$ controls the period of the quasi-oscillatory behavior of the FN model exhibited in Figure~\ref{fig:fitzdata}~\citep[e.g.,][]{rocsoreanu.etal12}, which is a prime example of when incorporating information from the observed data on the forward pass (\dalton) rather than the backward pass (\fenrir) is most useful. % At smaller step sizes, both \dalton and \fenrir converge to the true Laplace posterior. At the largest step size, \chk does not converge and requires $\dt \leq 0.05$ to cover all the true parameter values. \magi covers the true parameter values at all step sizes but notably has wider posteriors.
% is unable to cover the true $c$ unlike \dalton.

% \begin{figure}[!htb]
%     \includegraphics[width=\linewidth]{figures/fitzfigure.pdf}
%     \caption{
%         Parameter posteriors for the FN model using various ODE solvers.
%     }
%     \label{fig:fitzpost}
% \end{figure}

Table~\ref{tab:time} displays the timing for each of the probabilistic posterior estimates in Figure~\ref{fig:fitzpost}. The \dalton and \fenrir Laplace approximations are 1-2 orders of magnitude faster than either of the methods requiring MCMC. 
% total time spent at each step size for each probabilistic solver to produce parameter posteriors in Figure~\ref{fig:fitzpost}. As expected, \dalton and \fenrir are the fastest with comparable times since they both use the Laplace approximation while \chk and \magi require much longer to produce MCMC samples. 
\begin{table}[h]
    \caption{Time spent on the parameter inference results in seconds.} \label{tab:time}
    \begin{center}
        \begin{tabular}{@{} *{5}{c} @{}}
            \headercell{\\ Method} & \multicolumn{4}{c@{}}{Step Size ($\dt$)}\\
            \cmidrule(l){2-5}
            & 0.2 &  0.1 & 0.05 & 0.025   \\ 
            \midrule
            \dalton  & 15 &  16 &  17 &  16 \\
            \fenrir  & 16 &  15 &  15 &  17 \\
            \chk & - & 92 & 189 & 372 \\
            \magi & 222 & 418 & 742 & 1349
        \end{tabular}
    \end{center}
\end{table}

\subsection{SEIRAH model}\label{sec:seirah}

The SEIRAH model is a six-compartment epidemiological model used describe Covid-19 dynamics by~\cite{prague.etal20}. 
The six compartments of a given population are: susceptible ($S_t$), latent ($E_t$), ascertained infectious ($I_t$ ), removed ($R_t$), unascertained infectious ($A_t$), and hospitalized ($H_t$).  The SEIRAH model is
% .  The evolution of these quantities according to the SEIRAH model is given by
\begin{equation} \label{eq:seirah}
    \begin{aligned}
        \dot S_t &= -\frac{bS_t(I_t + \alpha A_t)}{N}, &
        \dot E_t &= \frac{bS_t(I_t + \alpha A_t)}{N} - \frac{E_t}{D_e}, \\
        \dot I_t &= \frac{rE_t}{D_e} - \frac{I_t}{D_q} - \frac{I_t}{D_I}, &
        \dot R_t &= \frac{I_t + A_t}{D_I} + \frac{H_t}{D_h}, \\
        \dot A_t &= \frac{(1-r)E_t}{D_e} - \frac{A_t}{D_I}, &
        \dot H_t &= \frac{I_t}{D_q} - \frac{H_t}{D_h},
    \end{aligned}
\end{equation}
where $N = S_t + E_t + I_t + R_t + A_t + H_t$ is fixed and so is $D_h=30$~\citep{prague.etal20}.  In this case ODE variables are both partially unobserved and subject to non-Gaussian measurement errors.
% Again we only have (indirect) observations of two of the six compartments.
That is, for
\begin{equation}
    \begin{aligned}\label{eq:seirahobs}
        \tilde I(t) &= \frac{rE_t}{D_e}, &
        \tilde H(t) &= \frac{I_t}{D_q},
    \end{aligned}
\end{equation}
% denote the new cases entering their respective compartments.  The measurement model of~\cite{prague.etal20} is
% Since the observed data are count processes, a reasonable noise model~\citep{prague.etal20} is
the measurement model is
\begin{equation}\label{eq:seirahnoise}
    \begin{aligned}
        Y_i^{(K)} \ind \operatorname{Poisson}(\tilde K(t_i)), \qquad K \in \{I, H\},
    \end{aligned}
\end{equation}
where $t_i = i$ days with $i = 0, \ldots, 60$.  The parameters to be estimated are $\tth = (b, r, \alpha, D_e, D_q, E_0, I_0)$, which we set to
% .  Data was simulated with true parameter values
$\tth = (2.23, 0.034, 0.55, 5.1, 2.3, 1.13, 15,492, 21,752)$.  The remaining ODE initial values are set to $S_0 = 63,884,630$, $R_0 = 0$, $A_0 = 618,013$, and $H_0 = 13,388$, as reported in~\cite{prague.etal20}. 

\begin{figure}[!htb]
  \includegraphics[width=\linewidth]{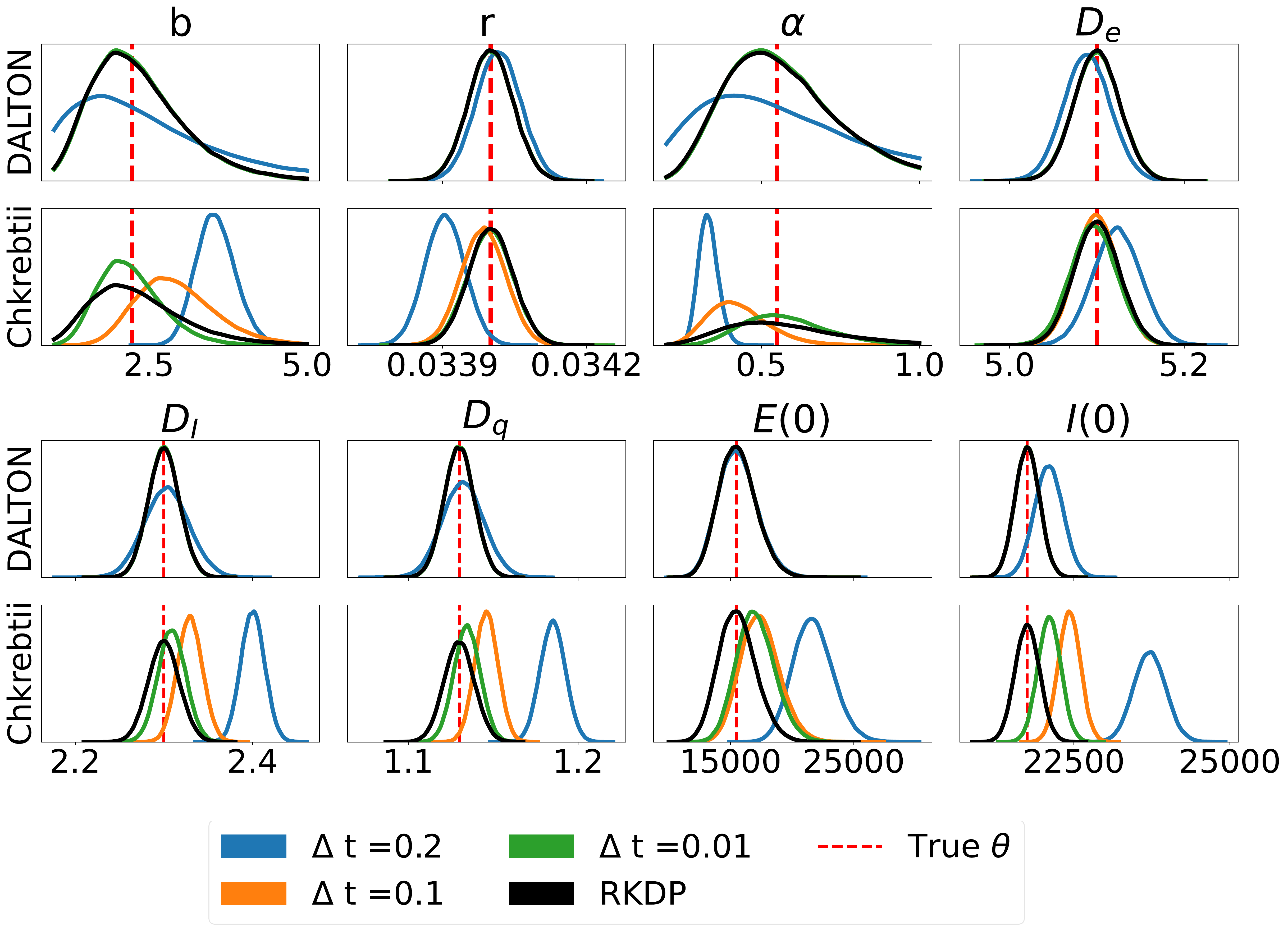}
  \caption{
    Parameter posteriors for the SEIRAH model for the \dalton and \chk methods.
  }
  \label{fig:seirahpost}
\end{figure}

Since the measurement model is non-Gaussian, \fenrir cannot be used.  Moreover, the \magi implementation provided by~\cite{wong2022magi} is restricted to Gaussian measurement models.  Therefore, Figure~\ref{fig:seirahpost} displays SEIRAH posterior estimates only for \dalton, \chk, and \rk.
% displays the \dalton and \chk posteriors for various step sizes $\dt$. Since observations are non-Gaussian, we use the non-Gaussian \dalton method described in~\ref{subsec:ngdalton} using the Algorithm~\ref{alg:daltonng} in Appendix~\appref{sec:ngalg}. The other two competing methods, \fenrir and \magi cannot be used in this example because the observation noise is non-Gaussian.
At all but the lowest stepsize, \dalton is indistinguishable from the Laplace approximation with the true likelihood (\rk).  In constrast, the \chk approximation has not converged to that of \rk even at the smallest step size $\dt = 0.05$.
% At the largest step size $\dt = 0.2$, \dalton covers all the true parameter values but \chk fails to do so for $D_I, D_q$ and $I(0)$. At smaller step sizes $\dt < 0.2$, \dalton converges to the true Laplace posterior.

\subsection{Fast Oscillation Model}\label{sec:osc}

The following second-order ODE is known for having many local maxima in its parameter likelihood function~\citep{tronarp22}: 
%A fast oscillations model known for having many local maxima in its parameter likelihood function~\citep{tronarp22} is a second-order ODE given by
\begin{equation} \label{eq:pen}
  \begin{aligned}
    \ddot x(t) &= -\frac{g}{L}\sin(x(t)),
  \end{aligned}
\end{equation}
where $\ddot x(t) = \dv[2]{t}x(t)$ and $g=9.81$ is the gravity constant. The model parameters are $\tth = (L, x(0), \dot x(0))$ with $L > 0$, which are to be learned from the measurement model
\begin{equation}\label{eq:pennoise}
    Y_i \ind \N(\dot x(t_i), \phi^2),
\end{equation}
where $t_i = i$, $i=0, 1, \ldots 10$, and $\phi^2 = 0.1$. (The extension of the Bayesian filtering model~\eqref{eq:bnf} is presented in Appendix~\ref{sec:higher}.)  
% Note that noisy observations are only observed on one component $\dot x(t_i)$.
Following~\cite{tronarp22}, the true parameter value is set to $\tth = (1, 0, \pi/2)$, and the initial value for each parameter learning algorithm (i.e., mode-finding for Laplace; otherwise MCMC) 
% the mode-finding algorithm for the Laplace approximation of \dalton, \fenrir, and \rk, and the MCMC algorithms of \chk and \magi --
is set to $\tth_0 = (5, 0, \pi/2)$ -- a poorly-chosen initial guess for $L$.
% the mode-finding algorithm of the \dalton, \fenrir, and \rk Laplace posteriors and of the MCMC algorithms  Data was simulated with true parameter values $\tth = (1, 0, \pi/2)$. Using the same settings as~\cite{tronarp22}, we choose $\tth_0 = (5, 0, \pi/2)$ as the initial guess.

\begin{figure}[!htb]
    \includegraphics[width=\linewidth]{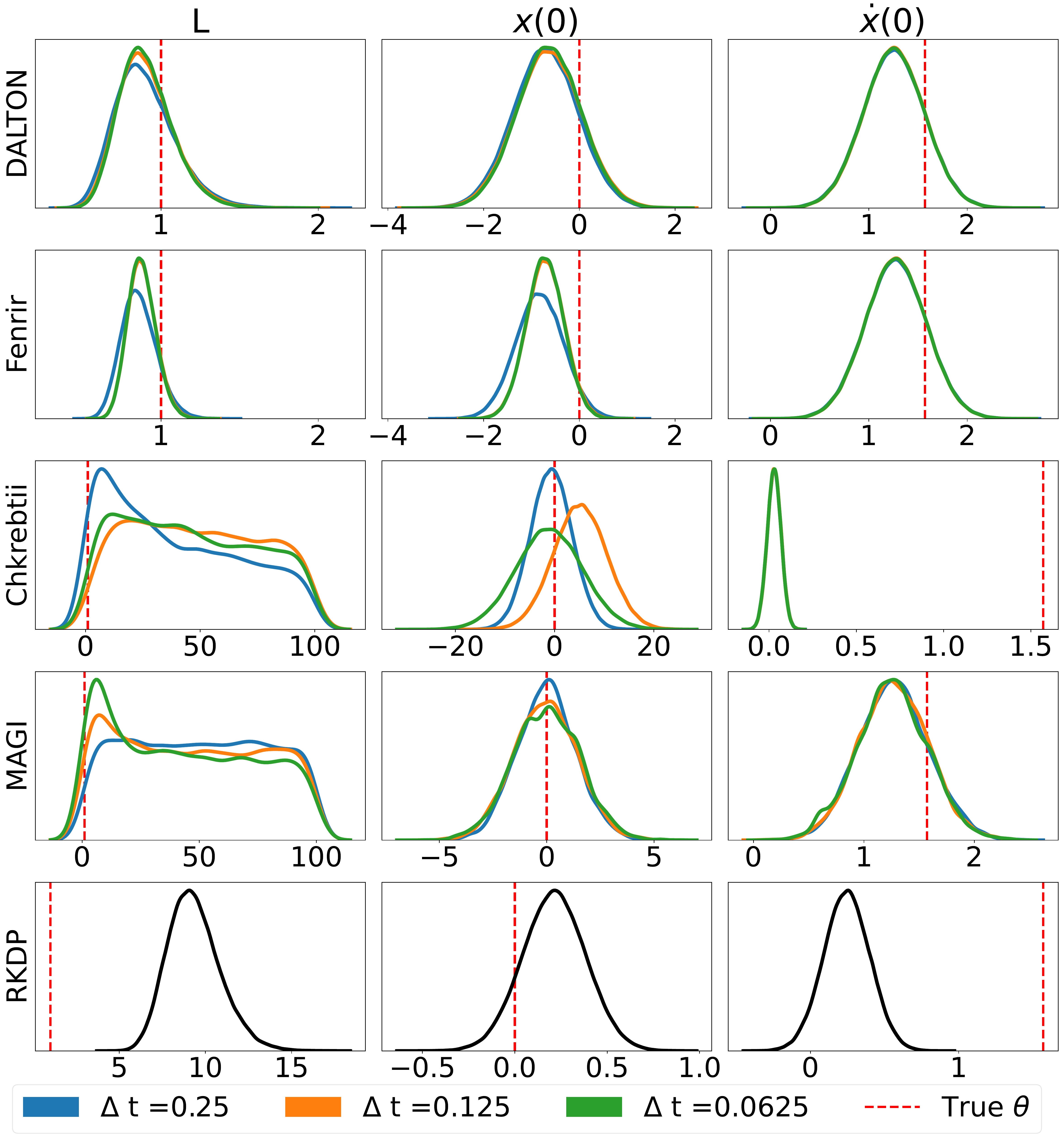}
    \caption{
        Parameter posteriors for the second-order ODE using various ODE solvers where the initial parameter value is $\tth_0 = (5, 0, \pi/2)$.
    }
    \label{fig:penpost}
\end{figure}

Figure~\ref{fig:penpost} displays posterior estimates for the four probabilistic likelihood approximations and for the true likelihood given by \rk.  \chk, \magi, and \rk provide very poor estimates of $L$.  In contrast, \dalton and \fenrir provide reasonable estimates of all parameters.  This experiment confirms the finding of~\cite{tronarp22} that mode-finding via \fenrir dominates other probabilistic likelihood approximations -- and even mode-finding with true likelihood itself -- when the latter is highly multimodal.  The same is now shown for mode-finding via \dalton.
% , the last of confirms the experiments of~\cite{tronarp22} which show how mode-finding on the true likelihood can get stuck in local modes.  In contrast, \dalton and \fenrir provide reasonable estimates of all three parameters. 
% Laplace posteriors for \dalton, \fenrir and \rk and the MCMC samples of \chk and \magi at different values of the solver step size $\dt = T/N$. \dalton and \fenrir cover the true parameter value at all step sizes. \chk fails to cover the true parameter value $L$ or $\dot x(0)$ depending on step size and \magi is unsure about $L$ with very wide posteriors. \rk fails to cover the parameter $L$ and $\dot x(0)$ which supports the results of~\cite{tronarp22}.

\subsection{Lorenz63 Model}\label{sec:lorenz}

% Having seen success with \dalton and \fenrir in the previous example, we now evaluate them on a chaotic system.
The Lorenz63 ODE~\citep{lorenz63} models the dynamics of a fluid layer induced by a warm temperature above and a cool temperature below. It is a classical example of a chaotic system, composed of three variables $\xx(t) = (x(t), y(t), z(t))$ satisfying
\begin{equation} \label{eq:lorenz63}
  \begin{aligned}
    \dot x(t) &= \alpha(y(t) - x(t)), \\
    \dot y(t) &= x(t)(\rho - z(t)) - y(t), \\
    \dot z(t) &= x(t)y(t) - \beta z(t).
  \end{aligned}
\end{equation}
The Lorenz63 model contains six parameters $\tth = (\rho, \alpha, \beta, x(0), y(0), z(0))$ with $\rho, \alpha, \beta > 0$. The measurement error model is
\begin{equation}\label{eq:lorenznoise}
    \YY_i \ind \N(\xx(t_i), \phi^2 \cdot \Id_{3\times 3}),
\end{equation}
where $t_i = i$, $i=0, 1, \ldots, 20$, and \mbox{$\phi^2 = 0.005$}.  The true parameter values are set to $\tth = (28, 10, 8/3, -12, -5, 38)$.

\begin{figure}[!htb]
    \centering
    \includegraphics[width=\linewidth]{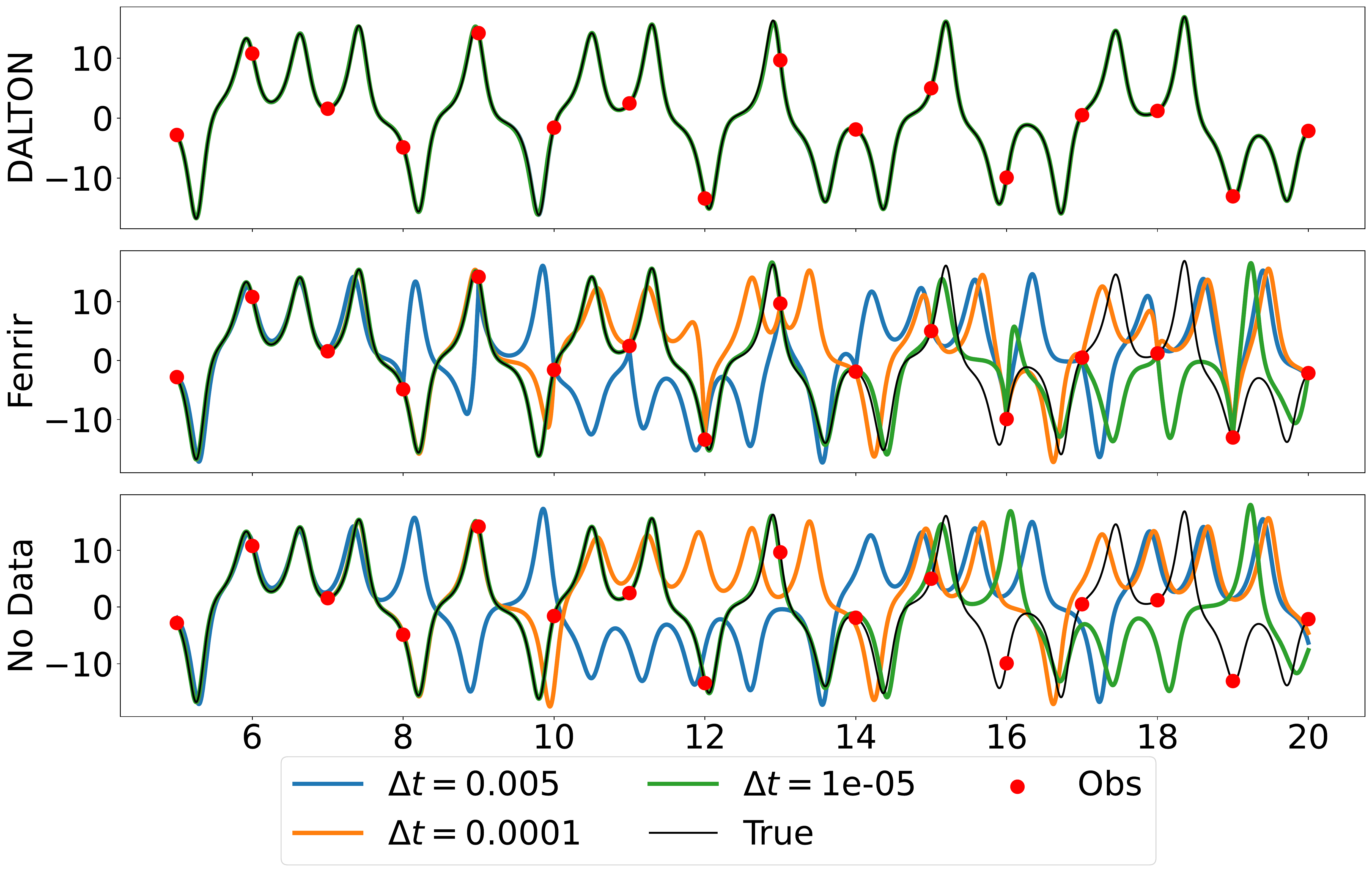}
    \caption{
        $x(t)$ for the Lorenz63 model calculated using various solvers. 
    }
    \label{fig:lorenzode}
\end{figure}

Figure~\ref{fig:lorenzode} displays various estimates of the system variable $x(t)$ assuming the model parameters are known (similar plots for $y(t)$ and $z(t)$ have been omitted for brevity).  Since Lorenz63 is even more sensitive to parameter values than the fast-oscillations model~\eqref{eq:pen}, \fenrir and \dalton are the only data-driven estimators considered.  Also plotted is the true solution given by \rk, and the data-free probabilistic solution using the linearization of~\eqref{eq:bnf} common to both \fenrir and \dalton (described in Appendix~\ref{sec:first}).  
% Before conducting parameter inference, we first estimate the ODE with the \fenrir and \dalton solvers and compare them to true ODE produced by \rk. For reference, we also estimate the ODE with the solvers assuming no data is given. The corresponding curves for $x(t)$ are displayed in Figure~\ref{fig:lorenzode}, with true parameter value $\tth = (28, 10, 8/3, -12, -5, 38)$ and at various step sizes. The curves for the $y(t)$ and $z(t)$ tell a similar story thus we omit them for brevity.
% and true parameter values b compares the stochastic ODE solutions produced by \dalton and \fenrir at $\dt=0.005$. The noisy observations are generated with $\tth = (28, 10, 8/3, -12, -5, 38)$ and the same $\tth$ is used for each probabilistic solver.
Like the data-free solver, \fenrir struggles with the chaotic nature of the Lorenz63 model even at very high resolution $(\dt = 10^{-5})$, being only able to incorporate information from the data on the backward pass. In contrast, \dalton effectively uses the observations on the forward pass to produce a solution close to the ground truth at much lower resolution.

\begin{figure}[!htb]
    \includegraphics[width=\linewidth]{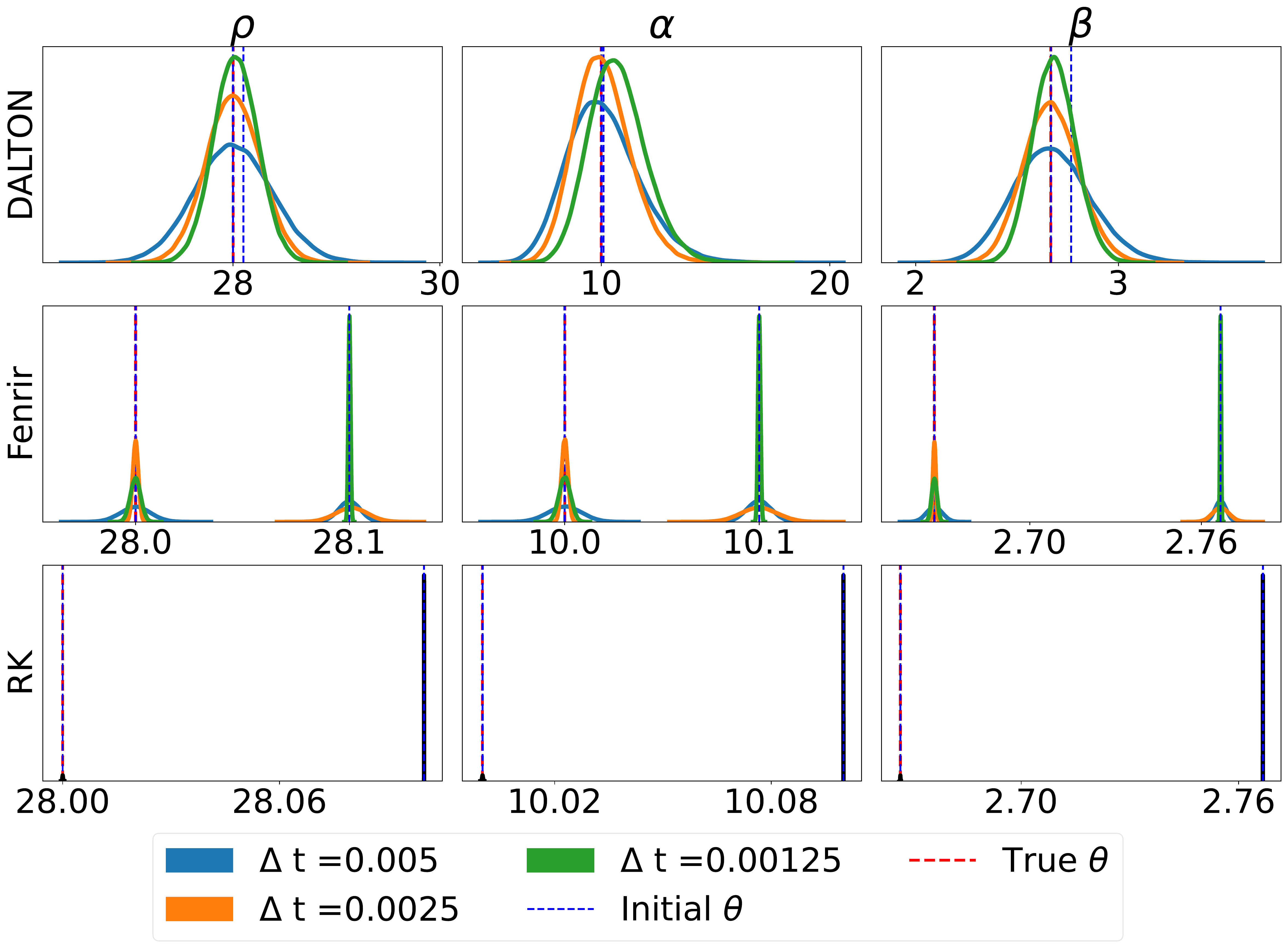}
    \caption{
        Parameter posteriors for the Lorenz63 model using various ODE solvers at two initial parameter values.
    }
    \label{fig:lorenzpost}
\end{figure}

Figure~\ref{fig:lorenzpost} displays the \dalton, \fenrir, and \rk posteriors, with mode-finding routines initialized at two different values of $\tth$: at and slightly away from the true parameter values.  When initialized at the true $\tth$, the \fenrir and \rk posteriors cover the true parameter values with vanishingly little uncertainty.  However, when their mode-finding algorithms are initialized slighly away from the true $\tth$, uncertainty remains small but the coverage drops to zero.  In contrast, the \dalton posteriors always cover the true parameter values and don't depend on the mode-finding algorithm's initial value.  This appears to be due to deep local optima in the \fenrir and \rk likelihoods which \dalton successfully removes. 

% with two initial $\tth$. The \dalton posteriors all cover the true parameter values. When the initial guess is the true parameter value, then both \fenrir and \rk posteriors do so as well, but with vanishingly little uncertainty.  However, when the initial $\tth$ is not the true parameter value, \fenrir and \rk were not able to escape the deep local maxima of their likelihoods. 
% and \fenrir parameter posteriors at $\dt = 0.005, 0.0025, 0.00125$. Also included are the parameter posteriors for \diffrax \rk. For \rk, and \fenrir, the posteriors are stuck at the initial values with very small variance. In contrast, \dalton is able to capture the uncertainty involved.

\begin{figure}[!htb]
    \centering
    \includegraphics[width=\linewidth]{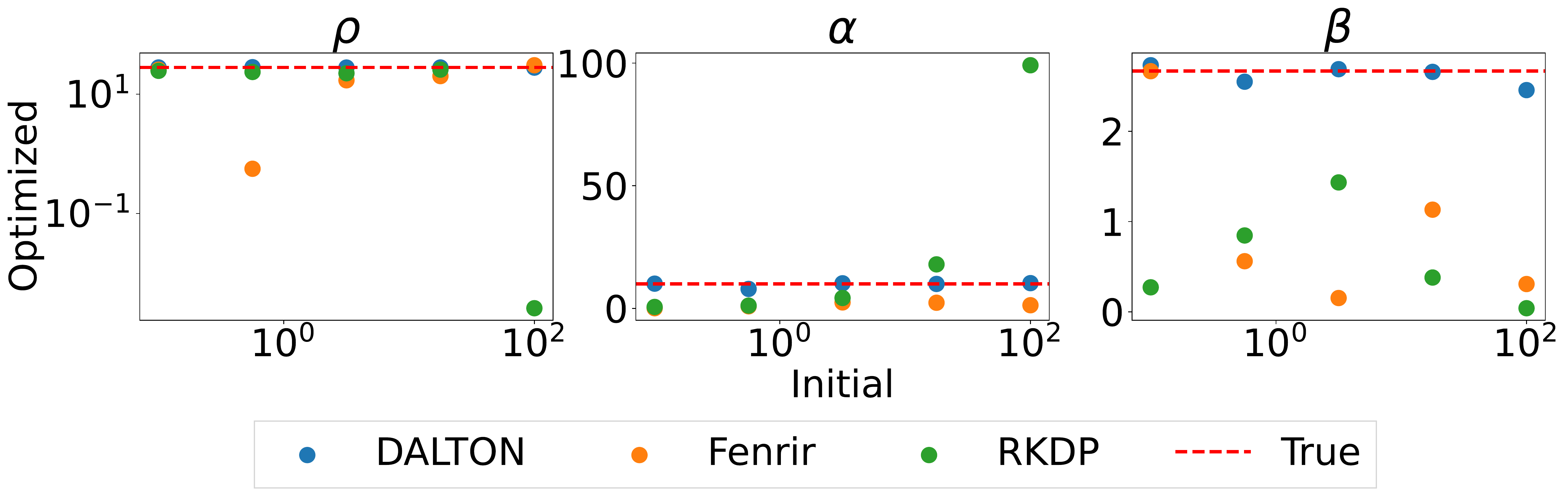}
    \caption{
        Initial and optimized parameter values for the Lorenz63 model using various ODE solvers.
    }
    \label{fig:lorenzip}
\end{figure}

To investigate this claim, we modify the parameter learning problem so that $t_i = i/10$ with $i=0,1,\ldots, 200$ and with initial values $(x(0), y(0), z(0))$ assumed to be known.  
% we now suppose that the measurement error model is the same as~\eqref{eq:lorenznoise} except $t_i = i/10$ with $i=0,1, \ldots, 200$. In other words, we now have $201$ observations instead of $21$. Also, we assume the initial values $x(0), y(0), z(0) = -12, -5, 38$ are known so that $\tth = (\alpha, \rho, \beta)$ are the only model parameters to estimate.
Figure~\ref{fig:lorenzip} displays the posterior modes of $\alpha$, $\rho$, and $\beta$ against different initializations of the mode-finding algorithms for \dalton, \fenrir and \rk. When initialized far from the ground truth, \fenrir and \rk fail to converge to the correct parameter values.  Since their posterior uncertainty in all cases is negligible, even the true Laplace posterior produced by \rk is unusable for parameter inference.  In contrast, \dalton is able to converge to the true parameters at a wide range of initial values, while retaining the parameter coverage exhibited in Figure~\ref{fig:lorenzpost}.

% \comment{I added Figure 5 on the log-scale for y-axis. You can decide which looks better. I could do a mixture of log and regular. For example log $\rho$ but keep regular $\alpha$ and $\beta$.}
% \begin{figure}[H]
%     \begin{center}
%         \includegraphics[width=\linewidth]{figures/lorenzip2.pdf}
%     \end{center}   
%     \caption{
%         Initial and optimized parameters for the lorenz63 model using \dalton, \fenrir and \diffrax \rk.
%     }
%     \label{fig:lorenzip2}
% \end{figure}

\section{Conclusion}\label{sec:conclusion}

We present \dalton, a probabilistic approximation to the intractable likelihood of ODE-IVP problems.  \dalton scales linearly in both time discretization points and ODE variables, accommodating both partially-observed components and non-Gaussian measurement models.  By incorporating information from the observed data in an online maner, \dalton can greatly reduce parameter sensitivity in many ODE systems.  It is thus an excellent candidate for parameter learning via mode-finding algorithms, outperforming existing probabilistic ODE likelihood approximations, and in cases of extreme parameter sensitivity, even the exact ODE likelihood itself.

% By incorporating information from the observed data in an online manner, \dalton can greatly reduce the sensitivity of many ODE systems to parameter values, thus providing more reliable parameter estimates than many other probabilistic ODE solvers, and in cases of extreme parameter hypersensitivity, than the exact ODE likelihood itself.  \dalton achieves this in a computationally competitive linear scaling regime by building on a well-established paradigm of Bayesian filtering which makes heavy use of the Kalman filtering and smoothing recursions. \dalton is performs well even on non-Gaussian partially unobserved measurements.

Despite empirical evidence of desirable performance, \dalton presently lacks theoretical convergence guarantees.  It should be noted, however, that the \magi and \fenrir methods do not have complete theoretical guarantees either: only maximum \emph{a posteriori} (MAP) convergence is known of the former~\citep{tronarp.etal21}, and forward-pass convergence of the latter~\citep{kersting20}.  Another limitation of \dalton is determining the appropriate step size.  A possible direction is to do this adaptively as described in~\cite{schober.etal19}, though it is unclear how the \dalton ratio terms in~\eqref{eq:condp} and~\eqref{eq:nongdec} will behave with a variable number of terms, nor how to efficiently backpropagate derivatives through the ODE parameters in that case. 
% One limitation of the \dalton solver is the determination the appropriate step size, which could possibly be done adaptively as described in~\cite{schober.etal19} but this is non-trivial. Another is the lack thereof theoretical guarantees of convergence common to many existing probabilistic ODE solvers~\citep{yang.etal21,tronarp22}.
There is also potential to explore the effectiveness of \dalton in estimating the parameters of stiff ODE systems, and to extend it to more complex boundary conditions.

% of parameter learning   which uses information from the observed data to 

% We present \dalton, a probabilistic method for parameter estimation that directly uses observations as a part of solution process. It has been demonstrated that \dalton can more effectively use observations compared to \fenrir to construct marginal likelihood approximations on several ODE parameter learning problems. One limitation of the \dalton solver is the determination the appropriate step size, which could be done adaptively as described in~\cite{schober.etal19}. 

%% file: supp.tex
\section{Higher Order ODE Systems} \label{sec:higher}

For a multi-variable function $\xx(t) = \big(x_1(t), \ldots, x_d(t)\big)$, an arbitrary-order ODE-IVP can be written as
\begin{equation}\label{eq:hode}
  \WW \XX(t) = \ff(\XX(t), t), \qquad \XX(0) = \vv, \qquad t \in [0, T],
\end{equation}
where $\XX(t) = (\XX^{(1)}(t), \ldots, \XX^{(d)}(t))$,
\begin{equation}\label{eq:xderivs}
  \XX^{(k)}(t) = \left(x_k(t), \dot x_k(t), \ldots, \dv[q_k-1]{t} x_k(t)\right)
\end{equation}
contains $x_k(t)$ and its first $q_k-1$ derivatives, $\WW$ is a coefficient matrix of size $r \times q$, where $q = \sum_{k=1}^d q_k$, and $\ff: \mathbb{R}^{q} \to \mathbb{R}^r$ is the (typically nonlinear) ODE function.
% are coefficient matrices with $\WW_k \in \mathbb{R}^{r_k \times q_k}$ and $\ff = \big(\ff_1, \ldots, \ff_d)$ are nonlinear functions with $\ff_k$ representing $r_k$ equations for $k=1,\ldots,d$.
For the usual ODE-IVP formulation of $\frac{\ud}{\ud t} \XX(t) = \ff(\XX(t), t)$, we have $\WW = \diag(\SS_{q_1}, \ldots, \SS_{q_d})$, where $\SS_{q_k} = [\bz_{(q_k-1)\times 1} \mid \Id_{(q_k-1) \times (q_k-1)}]$ for $k=1, \ldots, d$.  The Bayesian filtering model~\eqref{eq:bnf} in the data-free setting then becomes
\begin{equation}\label{eq:gbnf}
  \begin{aligned}
    \XX_{n+1} \mid \XX_n & \sim \N(\QQ \XX_n, \RR) \\
    \ZZ_n & \ind \N(\WW \XX_n - \ff(\XX_n, t_n), \VV),
  \end{aligned}
\end{equation}
with subsequent linearizations, data-driven extensions, etc., following straighforwardly from~\eqref{eq:gbnf}.

% Note that $\WW_\tth$ can be made to be block diagonal by padding $\WW_k$ with $p-q_k$ columns of zeros for $k=1, \ldots d$. This in combination with the linearization discussed in Appendix~\ref{sec:first}, allows for linear scaling in $d$ for higher-order ODE as well. Using this extension, \dalton can directly solve for a higher order ODE instead of the standard method of expressing it as multiple first order ODE employed by deterministic solvers, i.e., Runge-Kutta~\citep{dormand.prince80}.

\section{Gaussian Markov Process Prior} \label{sec:prior}

For the multi-variable function $\xx(t) = (x_1(t), \ldots, x_d(t))$, \dalton uses a simple and effective prior proposed by~\cite{schober.etal19}; namely, that each $x_k(t)$ follows an independent $(p_k-1)$-times integrated Brownian motion (IBM),
\begin{equation}\label{eq:ibm}
  \dv[p_k-1]{t} x_k(t) = \sigma_k B_k(t),
\end{equation}
where $B_1(t), \ldots, B_d(t)$ are independent Wiener processes.  This prior formulation gives $x_k(t)$ $p_k-1$ continuous derivatives.  It is therefore desirable to set $p_k > q_k$ to have a sufficiently smooth solution process.  In our experiments we have set $p_k = q_k + 1$, and redefined the ODE system\eqref{eq:hode} to be of order $q_k = p_k$ by padding $\WW$ with zeros.

The IBM prior~\eqref{eq:ibm} defines $\XX^{(k)}(t)$ in~\eqref{eq:xderivs} as a continuous Gaussian Markov process,
% Recall from~\eqrefmode that for a multi-variable function $\xx(t) = (x_1(t), \ldots, x_d(t))$, we define $\XX(t) = (\XX_1(t), \ldots, \XX_d(t))$ with $\XX_k(t) = (x_k(t), \dot x_k(t))$.
% % Recall from~\eqrefmode that for a multi-variable function $\xx(t) = (x_1(t), \ldots, x_d(t))$, we define $\XX(t) = (\XX_1(t), \ldots, \XX_d(t))$ with $\XX_k(t) = (x_k^{(0)}(t), \ldots, x_k^{(q_k-1)})$ containing the first $q_k-1$ derivatives of $x_k(t) = x_k^{(0)}(t)$.  Similarly, we define the initial value to be $\vv = (\vv_1, \ldots, \vv_d)$, and the coefficient matrix to be $\WW = (\WW_1, \ldots, \WW_d)$.
% \dalton uses a simple and effective prior proposed by~\cite{schober.etal19}; namely, that each $\XX_k(t)$, $k = 1,\ldots, d$ are independent 
% % $\tilde \XX_k(t) = \XX_k(t) - \vv_k$ is
% $p-1$ times integrated Brownian motion (IBM),
% \begin{equation}\label{eq:ibm}
%   x_k^{(p-1)}(t) = \sigma_k B_k(t),
% \end{equation}
% where $x_k^{(p-1)}(t)$ is the $p-1$-th derivative. 
% % where $p = \max{q_k} + 1$.
% Here we choose $p = 3$ because it is often advantageous in practice to set $p-1$ to be greater than the number of derivatives in the problem to increase the smoothness of $\XX_k(t)$ and at the same time keep $p$ small to reduce complexity.  This results in a $p$-dimensional continuous Gaussian Markov process on $\XX_k(t)$ defined by
\begin{equation}\label{eq:gmprm}
  \XX^{(k)}(t+\dt) \mid \XX^{(k)}(t) \sim \N(\QQ^{(k)}\XX^{(k)}(t), \RR^{(k)}),
\end{equation}
% $\XX_k(t) = \big(x_k^{(0)}(t), x_k^{(1)}(t), \ldots, x_k^{(p-1)}(t)\big)$
with the elements of matrices $\QQ^{(k)}$ and $\RR^{(k)}$ % in~\eqreflin
given by
\begin{equation}\label{eq:ibmprm}
  Q^{(k)}_{ij} = \mathbb{I}(i\leq j) \cdot \frac{(\dt)^{j-i}}{(j-i)!}, \qquad R^{(k)}_{ij} = \sigma_k^2 \cdot \frac{(\dt)^{2p-1-i-j}}{(2p-1-i-j)(p-1-i)!(p-1-j)!}.
\end{equation}
The full matrices in the Bayesian filtering model~\eqref{eq:gbnf} are
% We construct
$\QQ = \diag(\QQ^{(1)}, \ldots, \QQ^{(d)})$ and $\RR = \diag(\RR^{(1)}, \ldots, \RR^{(d)})$. % to be block diagonal.
In the context of parameter learning, the parameters $\eet$ of the prior process are the IBM scale parameters, $\eet = (\sigma_1, \ldots, \sigma_d)$.
% Furthermore, for each $k=1,\ldots, d$ we pad $\vv_k$ in~\eqrefmode with $p-2$ zeros.
% % and $p_k-q_k$ columns of zeros, respectively, to guarantee $\WW$ to be block diagonal. 
% A different method of padding $\vv_k$ is to work out the values of $x^{(l)}(0)$ for $2 \le l < p$ by taking derivatives of the ODE in~\eqrefmode.

\section{Efficient and Accurate ODE Linearization}\label{sec:first}

In the experiments of Section~\ref{sec:ex}, we employ a modified first-order linearization of $\ff(\XX_n, t_n)$ in~\eqref{eq:hode} proposed by~\citet{tronarp.etal18}.  That is, the first-order Taylor expansion of $\ff(\XX_n, t_n$ is
% also known as the extended Kalman filter (EKF) is
\begin{equation}\label{eq:first}
  \begin{aligned}
    \ff(\XX_n, t_n) \approx \ff(\mmu_{n|n-1}, t_n) + \JJ_n(\XX_n - \mmu_{n|n-1})
  \end{aligned}
\end{equation}
where $\mmu_{n|n-1}$ is the mean of the predictive distribution of the linearized working model~\eqref{eq:linpred} or~\eqref{eq:plindist}, depending on whether the linearization is data-free or data-dependent, 
% -- $p_\L(\XX_n \mid \ZZ_{0:n-1} = \bz)$ or $p_\L(\XX_n \mid \YY_{0:i(n-1)}, \ZZ_{0:n-1} = \bz)$ depending on whether the linearization is data-free or data-dependent --
and $\JJ_n = \pdv{\XX_n} \ff(\mmu_{n|n-1}, t_n)$.  The modified linearization is then
\begin{equation}
  \begin{aligned}
    \aa_n & = -\ff(\mmu_{n|n-1}, t_n) + \JJ_n \mmu_{n|n-1},
    & \BB_n & = \diag\left(\pdv{\XX^{(1)}_n} \ff(\mmu_{n|n-1}, t_n), \ldots, \pdv{\XX^{(d)}_n} \ff(\mmu_{n|n-1}, t_n)\right),
    & \VV_n & = \bz,
  \end{aligned}
\end{equation}
where $\XX_n = (\XX_n^{(1)}, \ldots, \XX_n^{(d)})$.  In other words, $\BB_n$ keeps only the block diagonal entries of $\JJ_n$.  With corresponding block diagonal matrices $\QQ_\eet$ and $\RR_\eet$, the Kalman recursions underlying the \fenrir and \dalton algorithms can be performed blockwise~\citep{kramer21}.  Thus for $q_k \equiv p$, the scaling of these algorithms drops from $\bO(N(dp)^3)$ to $\bO( Ndp^3 )$, a substantial improvement for low-order many-variable ODE systems.

% linearized working model $ = E_\L[\XX_n \mid \YY_{0:i(n-1)}, Z_{0:n} = \bz]$ is the predictive mean of the linearized model~\eqref{eq:plindist},

% the $\hat \xx_n = \E[\XX_n \mid \ZZ_{1:n-1}]$, and $J_f$ is the Jacobian of $\xx \to \ff(\xx, t)$. 

% For a multivariate ODE, a substantial computation speedup from $\bO(N(dp)^3)$ to $\bO( Ndp^3 )$ in the Kalman filter is to assume each matrix in~\eqreflin is block diagonal~\citep{kramer21}. This can be done by assuming the priors for each variable is independent, such that, the prior matrices, $\QQ_\eet$ and $\RR_\eet$ are block diagonal~\citep{kramer21}. The only remaining matrices in~\eqreflin are $\BB_n$ and $\VV_n$. For the zeroth order Taylor approximation, both $\BB_n = \VV_n = \bz$. In the first order case, the Jacobian, $J_f$, is not necessarily block diagonal. However, it is noted that keeping only the block diagonal elements of $J_f$ only slightly reduces numerical accuracy~\citep{kramer21}. Denoting the block diagonal of $J_f$ as $\bd(J_f)$, and setting $\aa_n = -\ff(\hat \xx_n) + \bd(J_f)(\hat \xx_n)\hat \xx_n$, $\BB_n = - \bd(J_f)(\hat \xx_n)$ and $\VV_n = 0$ ensures all matrices in~\eqreflin are block diagonal.

\section{Implementation Details for Parameter Learning Methods}\label{sec:parpost}

In the experiments of Section~\ref{sec:ex}, We take the measurement model parameters $\pph$ to be known, such that the learning problem is only for the ODE model parameters $\tth = (\theta_1, \ldots, \theta_D)$ and the tuning parameter $\eet = (\sigma_1, \ldots, \sigma_d)$ of the IBM prior in Appendix~\ref{sec:prior}.  We use a flat prior on $\eet$ and independent $\N(0, 10^2)$ priors on either $\theta_r$ or $\log \theta_r$, $r = 1, \ldots, D$, depending on whether $\theta_r$ is unbounded or $\theta_r > 0$.

For the unknown model parameters $\tth$, we use the Laplace posterior approximation
\begin{equation}\label{eq:bnaapp}
  \tth \mid \YY_{0:M} \approx \N(\hat \tth, \hat \VV_{\tth}),
\end{equation}
where $(\hat \tth, \hat \eet) = \argmax_{(\tth, \eet)} \log p(\tth, \eet \mid \YY_{0:M})$ is obtained using the Newton-CG optimization algorithm~\citep{NoceWrig06} as implemented in the \python \textbf{JAXopt} library~\citep{jaxopt}.  As for $\hat \VV_{\tth}$, we estimate it as
% and $\hat \VV_{(\tth, \eet)} = -\big[\pdv{}{\TTh}{\TTh'} \log p(\hat \TTh \mid \YY_{0:M})\big]^{-1}$. To obtain $\hat \TTh$ we use the Newton-CG optimization algorithm~\citep{NoceWrig06} as implemented in the \textbf{JAXopt} library~\citep{jaxopt}.  The Laplace approximation $p(\tth \mid \YY_{0:M}) \approx \N(\hat \tth, \hat \VV_{\tth})$ was obtained from $(\hat \tth, \hat \eet) = \argmax_{(\tth, \eet)} \log p(\TTh \mid \YY_{0:M})$ and
$\hat \VV_{\tth} = -\big[\pdv{}{\tth}{\tth'} \log p(\hat \TTh \mid \YY_{0:M})\big]^{-1}$, i.e., with the Hessian taken only with respect to $\tth$ and not $\eet$.  This approach was used in~\cite{tronarp22}, and found here to produce slightly better results than when uncertainty is propagated through the prior process tuning parameters as well.

The \magi algorithm as implemented in~\cite{wong2022magi} uses a slighly different prior specification, i.e., with a Mat{\'e}rn covariance function with smoothing parameter $\nu = 2.01$, which ensures twice-continuous differentiability of the prior on $\xx(t)$, and scale parameters $\sigma_k$ corresponding to the maximum \emph{a posteriori} (MAP) estimates of the Gaussian process regression conditioning only on the observed data, which is analytically tractable since this implementation of \magi is restricted to Gaussian measurement models.  The implementation then uses a hybrid Monte Carlo (HMC) algorithm to sample from $p(\tth, \XX_{0:N} \mid \YY_{0:M})$.  

Our implementation of the \chk MCMC algorithm starts by fitting a \dalton parameter learning algorithm, which we do not factor into the timings presented in Table~\ref{tab:time}.  From this preliminary inference run we fix the Gaussian process tuning parameters at $\hat \eet$.  The \chk algorithm still requires the specification of a proposal distribution $p(\tth \mid \tth')$.  We set this as $\tth \sim \N(\tth', \tau \cdot \hat \VV_{\tth})$, where $\hat \VV_{\tth}$ is obtained from the \dalton algorithm and $\tau$ is tuned during the burning stage to obtain an overall acceptance rate of 25\% (also not counted in the timings of Table~\ref{tab:time}).

\newpage

\section{\dalton Algorithms}\label{sec:alg}

For Algorithms~\ref{alg:dalton} and~\ref{alg:daltonng}, we use the extension discussed in Appendix~\ref{sec:higher} to generalize to arbitrary-order ODEs. The $\linearize()$ function (e.g., Line~\ref{ln:lin} of Algorithm~\ref{alg:dalton}) performs the modified first-order linearization detailed in Appendix~\ref{sec:first}, and $\nlogpdf()$ on Line~\ref{ln:logpdf} is the log-pdf of a multivariate normal.

\subsection{Gaussian Measurements}

\begin{algorithm}[!htb]
  \caption{\dalton probabilistic ODE likelihood approximation for Gaussian measurements.}\label{alg:dalton}
  \begin{algorithmic}[1]
    \Procedure{\code{dalton}}{$\WW_\tth, \ff(\XX, t, \tth), \vv_\tth, \QQ_\eet, \RR_\eet, \YY_{0:M}, \DD^\pph_{0:M}, \OOm^\pph_{0:M}$}
    \State $\mmu_{0|0}, \Sigma_{0|0} \gets \vv, \bz$ \Comment{Initialization}
    \State $\ZZ_{0:N} \gets \bz$
    \State $\ell_z, \ell_{yz} \gets 0, 0$
    \State $i \gets 0$ \Comment{Used to map $t_n$ to $t'_i$}
    \State \Comment{Lines 7-12 compute $\log p(\ZZ_{0:N} = \bz \mid \TTh)$}
    \For{$n=1:N$}
    \State $\mmu_{n|n-1}, \SSi_{n|n-1} \gets \kpredict(\mmu_{n-1|n-1}, \SSi_{n-1|n-1}, \bz, \QQ_\eet, \RR_\eet)$
    \State $\aa_n, \BB_n, \VV_n \gets \linearize(\mmu_{n|n-1}, \SSi_{n|n-1}, \WW_\tth, \ff(\XX, t_n, \tth))$ \label{ln:lin}
    \State $\mmu_n, \SSi_n \gets \kforecast(\mmu_{n|n-1}, \SSi_{n|n-1}, \aa_n, \WW_\tth + \BB_n, \VV_n)$
    \State $\ell_z \gets \ell_z + \nlogpdf(\ZZ_n = \bz; \mmu_n, \SSi_n)$ \label{ln:logpdf}
    \State $\mmu_{n|n}, \SSi_{n|n} \gets \kupdate(\mmu_{n|n-1}, \SSi_{n|n-1}, \ZZ_n, \aa_n, \WW_\tth + \BB_n, \VV_n)$ 
    \EndFor
    \State \Comment{Lines 14-24 compute $\log p(\YY_{0:M}, \ZZ_{0:N} = \bz \mid \TTh)$}
    \State $\ell_{yz} \gets \nlogpdf(\YY_0; \DD^\pph_0\vv_\tth, \OOm^\pph_0)$
    \For{$n=1:N$}
    \State $\mmu_{n|n-1}, \SSi_{n|n-1} \gets \kpredict(\mmu_{n-1|n-1}, \SSi_{n-1|n-1}, \bz, \QQ_\eet, \RR_\eet)$
    \State $\aa_n, \BB_n, \VV_n \gets \linearize(\mmu_{n|n-1}, \SSi_{n|n-1}, \WW_\tth, \ff(\XX, t_n, \tth))$
    \If{$t_n = t_{n(i)}$}
    \State $\ZZ_n \gets
    \begin{bmatrix}
      \ZZ_n \\
      \YY_i
    \end{bmatrix}, \qquad
    \WW_\tth \gets
    \begin{bmatrix} 
      \WW_\tth \\
      \DD^\pph_i
    \end{bmatrix}, \qquad
    \BB_n \gets
    \begin{bmatrix}
      \BB_n \\
      \bz
    \end{bmatrix}$
    \State $\aa_n \gets
    \begin{bmatrix}
      \aa_n \\
      \bz
    \end{bmatrix}, \qquad
    \VV_n \gets
    \begin{bmatrix}
      \VV_n & \bz \\
      \bz & \OOm^\phi_i
    \end{bmatrix}$ 
    \State $i \gets i + 1$
    \EndIf
    \State $\mmu_n, \SSi_n \gets \kforecast(\mmu_{n|n-1}, \SSi_{n|n-1}, \aa_n, \WW_\tth +\BB_n, \VV_n)$    
    \State $\ell_{yz} \gets \ell_{yz} + \nlogpdf(\ZZ_n; \mmu_n, \SSi_n)$
    \State $\mmu_{n|n}, \SSi_{n|n} \gets \kupdate(\mmu_{n|n-1}, \SSi_{n|n-1}, \ZZ_n, \aa_n, \WW_\tth + \BB_n, \VV_n)$ 
    \EndFor
    \State
    \State \textbf{return} $\ell_{yz} - \ell_z$ \Comment{Estimate of $\log p(\YY_{0:M} \mid \ZZ_{0:N} = \bz, \TTh)$}
    \EndProcedure
  \end{algorithmic}
\end{algorithm}

% \newpage

\subsection{Non-Gaussian Measurements}\label{sec:ngalg}

Recall in Section~\ref{subsec:ngdalton} that we denote $\xobs_{n(i)}$ and $\xun_{n(i)}$ to be the observed and unobserved components of $\xx(t)$ at time $t = t_i'$.  In order to present Algorithm~\ref{alg:daltonng}, we introduce the notation $\xobs_{n(i)} = \DD_i \XX_{n(i)}$, where $\DD_i$ is a mask matrix of zeros and ones, so as to more naturally relate to the Gaussian version of \dalton in Algorithm~\ref{alg:dalton}.
% In particular, the zero elements of $\DD_i$ are precisely those in which the Hessian matrix $\pdv{}{\XX_{n(i)}}{\XX_{n(i)}'} g_i(\YY_i, \XX_{n(i)}, \pph)$ 
% we use the Hessian of $g_i(\YY_i, \XX_{n(i)}, \pph)$ to find a mask matrix $\DD_i$ to match the equivalent $\DD^{\pph}_i$ in Gaussian \dalton. We do this by noting that the Hessian will be equal to $0$ for any unobserved components in Line~\ref{line:mask}. Using this mask matrix, we have $\xobs_{n(i)} = \DD_i \XX_{n(i)}$.
%\setcounter{algorithm}{1}
\begin{algorithm}[!htb]
  \caption{\dalton probabilistic ODE likelihood approximation for non-Gaussian measurements.}\label{alg:daltonng}
  \begin{algorithmic}[1]
    \Procedure{\code{dalton}}{$\WW_\tth, \ff(\XX, t, \tth), \vv_\tth, \QQ_\eet, \RR_\eet, \YY_{0:M}, \pph, \gg_{0:M}(\YY, \xobs, \pph)$}
    \State $\mmu_{0|0}, \Sigma_{0|0} \gets \vv, \bz$ \Comment{Initialization}
    \State $\ZZ_{0:N} \gets \bz$
    \State $\ell_{xz}, \ell_{xyz} \ell_y \gets 0, 0, 0$
    \State $i \gets 0$ \Comment{Used to map $t_n$ to $t'_i$}
    \State \Comment{Lines 7-24 compute $\log p(\XX_{0:N} \mid \hat \YY_{0:M}, \ZZ_{0:N}=\bz, \TTh)$}
    \For{$n=1:N$} 
    \State $\mmu_{n|n-1}, \SSi_{n|n-1} \gets \kpredict(\mmu_{n-1|n-1}, \SSi_{n-1|n-1}, \bz, \QQ_\eet, \RR_\eet)$
    \State $\aa_n, \BB_n, \VV_n \gets \linearize(\mmu_{n|n-1}, \SSi_{n|n-1}, \WW_\tth, \ff(\XX, t_n, \tth))$
    \If{$t_n = t_{n(i)}$}
%    \State $\hat \xx_n \gets \subsetfn(\mmu_{n-1|n-1}, t_n)$
%    \State $g \gets g_i(\hat \xx_n, \YY_i, \pph)$
%    \State $\gg_{(1)} \gets \nabla g_i(\hat \xx_n, \YY_i, \pph)$
%    \State $\gg_{(2)} \gets \nabla^2 g_i(\hat \xx_n, \YY_i, \pph)$
%    \State $\hat \YY_i \gets \hat \xx_n - \gg_{(2)}^{-1}\gg_{(1)}$ \Comment{Compute pseudo-observations}
    % \State $g \gets g_i(\YY_i, \mmu_{n|n-1}, \pph)$
    \State $\gg_{(1)} \gets \pdv{\xobs_{n(i)}} g_i(\YY_i, \DD_i \mmu_{n|n-1}, \pph)$
    \State $\gg_{(2)} \gets \pdv{}{\xobs_{n(i)}}{\xobs_{n(i)}'} g_i(\YY_i, \DD_i \mmu_{n|n-1}, \pph)$
    % \State $\DD_i \gets \code{where}(\gg_{(2)} \neq 0)$ 
    % \Comment{Mask matrix for which elements in Hessian is not 0} \label{line:mask}
    \State $\hat \YY_i \gets \DD_i \mmu_{n|n-1}- \gg_{(2)}^{-1}\gg_{(1)}$ \Comment{Compute pseudo-observations}
    \State $\ZZ_n \gets
    \begin{bmatrix}
      \ZZ_n \\
      \hat \YY_i
    \end{bmatrix}, \qquad
    \WW_\tth \gets
    \begin{bmatrix} 
      \WW_\tth \\
      \DD_i
    \end{bmatrix}, \qquad
    \BB_n \gets
    \begin{bmatrix}
      \BB_n \\
      \bz
    \end{bmatrix}$
    \State $\aa_n \gets
    \begin{bmatrix}
      \aa_n \\
      \bz
    \end{bmatrix}, \qquad
    \VV_n \gets
    \begin{bmatrix}
      \VV_n & \bz \\
      \bz & \gg_{(2)}^{-1}
    \end{bmatrix}$ 
    \State $i \gets i + 1$
    \EndIf 
    \State $\mmu_{n|n}, \SSi_{n|n} \gets \kupdate(\mmu_{n|n-1}, \SSi_{n|n-1}, \ZZ_n, \aa_n, \WW_\tth +\BB_n, \VV_n)$ 
    \EndFor
    \For{$n=N-1:1$} 
    \State $\mmu_{n|N}, \SSi_{n|N} \gets \ksmooth(\mmu_{n+1|N}, \SSi_{n+1|N}, \mmu_{n|n}, \SSi_{n|n}, \mmu_{n+1|n}, \SSi_{n+1|n}, \QQ_\eet)$
    \State $\mmu_n, \SSi_n \gets \ksample(\mmu_{n+1|N}, \mmu_{n|n}, \SSi_{n|n}, \mmu_{n+1|n}, \SSi_{n+1|n}, \QQ_\eet)$
    \State $\ell_{xyz} \gets \ell_{xyz} + \nlogpdf(\mmu_{n|N}; \mmu_n, \SSi_n)$
    \EndFor
    \State $\ell_{xyz} \gets \ell_{xyz} + \nlogpdf(\mmu_{N|N}; \mmu_{N|N}, \SSi_{N|N})$
    \State \Comment{Lines 26-35 compute $\log p(\XX_{0:N} \mid \ZZ_{0:N}=\bz, \TTh)$}
    \State $\ZZ_{0:N} \gets \bz$ \Comment{Reset $\ZZ_{0:N} =\bz$}
    \State $\mmu'_{0:N|N} \gets \mmu_{0:N|N}$ \Comment{Store $\mmu_{0:N|N} = \E[\XX_{0:N} \mid \ZZ_{0:N}, \hat \YY_{0:N}]$ for later}
    
    \For{$n=1:N$}
    \State $\mmu_{n|n-1}, \SSi_{n|n-1} \gets \kpredict(\mmu_{n-1|n-1}, \SSi_{n-1|n-1}, \bz, \QQ_\eet, \RR_\eet)$
    \State $\aa_n, \BB_n, \VV_n \gets \linearize(\mmu_{n|n-1}, \SSi_{n|n-1}, \WW_\tth, \ff(\XX, t_n, \tth))$
    \State $\mmu_{n|n}, \SSi_{n|n} \gets \kupdate(\mmu_{n|n-1}, \SSi_{n|n-1}, \ZZ_n, \aa_n, \WW_\tth + \BB_n, \VV_n)$ 
    \EndFor
    \For{$n=N-1:1$} 
    \State $\mmu_n, \SSi_n \gets \ksample(\mmu'_{n+1|N}, \mmu_{n|n}, \SSi_{n|n}, \mmu_{n+1|n}, \SSi_{n+1|n}, \QQ_\eet)$
    \State $\ell_{xz} \gets \ell_{xz} + \nlogpdf(\mmu'_{n|N}; \mmu_n, \SSi_n)$
    \EndFor
    \State $\ell_{xz} \gets \ell_{xz} + \nlogpdf(\mmu'_{N|N}; \mmu_{N|N}, \SSi_{N|N})$
    \State \Comment{Lines 37-39 compute $\log p(\YY_{0:M} \mid \XX_{0:N}, \pph)$}
    \For{$i=0:M$} 
%    \State $\xx_i \gets \subsetfn(\mmu'_{n(i)|N}, t_{n(i
%    \State $\ell_y \gets \ell_y - g_i(\xx_i, \YY_i, \pph)$
    \State $\ell_y \gets \ell_y - g_i(\YY_i, \mmu'_{n(i)|N}, \pph)$
    \EndFor
    \State 
    \State \textbf{return} $\ell_{xz} - \ell_y - \ell_{xyz}$ \Comment{Estimate of $\log p(\YY_{0:M} \mid \ZZ_{0:N} = \bz, \TTh)$}
    \EndProcedure
  \end{algorithmic}
\end{algorithm}

\newpage

\section{Kalman Functions}\label{sec:kalmanfun}

The Kalman recursions used in Algorithms~\ref{alg:dalton} and~\ref{alg:daltonng} are formulated in terms of the general Gaussian state space model
\begin{equation}
  \begin{aligned}
    \XX_n &= \QQ_n \XX_{n-1} + \cc_n + \RR_n^{1/2} \eps_n \\
    \ZZ_n &= \WW_n \XX_n + \aa_n + \VV_n^{1/2} \eet_n
  \end{aligned}, \qquad \eps_n, \eet_n \ind \N(\bz,\Id).
\end{equation}
Throughout, we use the notation $\mmu_{n|m} = \E[\XX_n \mid \ZZ_{0:m}]$ and $\SSi_{n|m} = \var(\XX_n \mid \ZZ_{0:m})$.
\begin{algorithm}[!htb]
  \caption{Standard Kalman filtering and smoothing recursions.}\label{alg:kalman}
  \begin{algorithmic}[1]
    % \LeftComment{Perform the prediction step.}
    \Procedure{\kpredict}{$\mmu_{n-1|n-1}, \SSi_{n-1|n-1}, \cc_n, \QQ_n, \RR_n$} % \Comment{Perform the prediction step.}
    \State $\mmu_{n|n-1} \gets \QQ_n\mmu_{n-1|n-1} + \cc_n$
    \State $\SSi_{n|n-1} \gets \QQ_n\SSi_{n-1|n-1}\QQ_n' + \RR_n$
    \State \textbf{return} $\mmu_{n|n-1}, \SSi_{n|n-1}$ % \Comment{Predicted mean and variance: $\E[\XX_n \mid \ZZ_{0:n-1}], \var[\XX_n \mid \ZZ_{0:n-1}]$}
    \EndProcedure
    % \LeftComment{Perform the update step.}
    \Procedure{\kupdate}{$\mmu_{n|n-1}, \SSi_{n|n-1}, \ZZ_n, \aa_n, \WW_n, \VV_n$} % \Comment{Perform the update step.}
    \State $\AA_n \gets \SSi_{n|n-1} \WW_n'[\WW_n\SSi_{n|n-1} \WW_n' + \VV_n]^{-1}$
    \State $\mmu_{n|n} \gets \mmu_{n|n-1}+\AA_n(\ZZ_n - \WW_n\mmu_{n|n-1} - \aa_n)$
    \State $\SSi_{n|n} \gets \SSi_{n|n-1} - \AA_n \WW_n\SSi_{n|n-1}$
    \State \textbf{return} $\mmu_{n|n}, \SSi_{n|n}$ % \Comment{Updated mean and variance: $\E[\XX_n \mid \ZZ_{0:n}], \var[\XX_n \mid \ZZ_{0:n}]$}
    \EndProcedure
    % \LeftComment{Perform the smoothing step.}
    \Procedure{\ksmooth}{$\mmu_{n+1|N}, \SSi_{n+1|N}, \mmu_{n|n}, \SSi_{n|n}, \mmu_{n+1|n}, \SSi_{n+1|n}, \QQ_{n+1}$} % \Comment{Perform the smoothing step.}
    \State $\AA_n \gets \SSi_{n|n}\QQ_{n+1}'\SSi_{n+1|n}^{-1}$
    \State $\mmu_{n|N} \gets \mmu_{n|n} + \AA_n(\mmu_{n+1|N} - \mmu_{n+1|n})$
    \State $\SSi_{n|N} \gets \SSi_{n|n} + \AA_n(\SSi_{n+1|N} - \SSi_{n+1|n}) \AA_n'$
    \State \textbf{return} $\mmu_{n|N}, \SSi_{n|N}$ % \Comment{Smoothing mean and variance: $\E[\XX_n \mid \ZZ_{0:N}], \var[\XX_n \mid \ZZ_{0:N}]$}
    \EndProcedure
    % \LeftComment{Perform the sampler smoothing step.}
    \Procedure{\ksample}{$\xx_{n+1}, \mmu_{n|n}, \SSi_{n|n}, \mmu_{n+1|n}, \SSi_{n+1|n}, \QQ_{n+1}$} % \Comment{Perform the sampler smoothing step.}
    \State $\AA_n \gets \SSi_{n|n}\QQ_{n+1}'\SSi_{n+1|n}^{-1}$
    \State $\tilde{\mmu}_{n|N} \gets \mmu_{n|n} + \AA_n(\xx_{n+1} - \mmu_{n+1|n})$ \Comment{$\tilde{\mmu}_{n|N} = \E[\XX_n \mid \XX_{n+1:N}, \ZZ_{0:N}]$}
    \State $\tilde{\SSi}_{n|N} \gets \SSi_{n|n} - \AA_n\QQ_{n+1}\SSi_{n|n}$ \Comment{$\tilde{\SSi}_{n|N} = \var[\XX_n \mid \XX_{n+1:N}, \ZZ_{0:N}]$}
    \State \textbf{return} $\tilde{\mmu}_{n|N}, \tilde{\SSi}_{n|N}$ % \Comment{Smoothing sampler mean and variance: $\E[\XX_n \mid \XX_{n+1}, \ZZ_{0:N}], \var[\XX_n \mid \XX_{n+1}, \ZZ_{0:N}]$}
    \EndProcedure
    % \LeftComment{Perform the forecast step.}
    \Procedure{\kforecast}{$\mmu_{n|n-1}, \SSi_{n|n-1}, \aa_n, \WW_n, \VV_n$} % \Comment{Perform the forecast step.}
    \State $\lla_n \gets \WW_n \mmu_{n|n-1} + \aa_n$ \Comment{$\lla_{n|n-1} = \E[\ZZ_n \mid \ZZ_{0:n-1}]$}
    \State $\OOm_{n|n-1} \gets \WW_n \mmu_{n|n-1} \WW_n' + \VV_n$ \Comment{$\OOm_{n|n-1} = \var[\ZZ_n \mid \ZZ_{0:n-1}]$}
    \State \textbf{return} $\lla_{n|n-1}, \OOm_{n|n-1}$ % \Comment{Forecasted mean and variance: $\E[\ZZ_n \mid \ZZ_{0:n-1}], \var[\ZZ_n \mid \ZZ_{0:n-1}]$}
    \EndProcedure
  \end{algorithmic}
\end{algorithm}